\documentclass[final,3p,authoryear,times,twocolumn]{elsarticle}

\usepackage{framed,multirow}

\usepackage{amssymb}
\usepackage{latexsym}

\usepackage{url}
\usepackage{xcolor}
\usepackage{tabu}

\usepackage{graphicx}
\usepackage{color}
\usepackage{amsmath}
\usepackage{threeparttable}

\usepackage{amssymb}
\usepackage[noadjust]{cite}
\usepackage{filecontents}

\DeclareMathOperator*{\argmax}{arg\,max}
\usepackage{multicol}
\usepackage{fancyhdr}

\usepackage{colortbl}
\usepackage{booktabs,xcolor,siunitx}
\usepackage{mathtools}
\usepackage{arydshln}
\definecolor{light-gray}{gray}{0.95}
\newcolumntype{L}[1]{>{\raggedright\let\newline\\\arraybackslash\hspace{0pt}}m{#1}}
\newcolumntype{C}[1]{>{\centering\let\newline\\\arraybackslash\hspace{0pt}}m{#1}}
\newcolumntype{R}[1]{>{\raggedleft\let\newline\\\arraybackslash\hspace{0pt}}m{#1}}

\usepackage{hyperref}
\usepackage{cuted}
\usepackage{lipsum}
\usepackage[percent]{overpic}

\usepackage[labelfont=bf,justification=raggedright,singlelinecheck=false]{caption}
\definecolor{newcolor}{rgb}{.8,.349,.1}

\newif\ifproofread
\newcommand{\changemarker}[1]{%
\ifproofread
\textcolor{black}{#1} 
\else
#1%
\fi
}
\proofreadtrue
\newif\ifrevreference

\journal{Medical Image Analysis}

\begin{document}
\revreferencefalse 

\begin{frontmatter}

\title{Robust Classification from Noisy Labels: Integrating Additional Knowledge for Chest Radiography Abnormality Assessment}%

\author[1,3]{Sebastian G\"undel\corref{cor1}}
\cortext[cor1]{Corresponding author at: Digital Technology and Inovation, Siemens Healthineers, 91052 Erlangen, Germany and Pattern Recognition Lab, Friedrich-Alexander-Universit\"at Erlangen-N\"urnberg, 91058 Erlangen, Germany \newline E-mail address: sebastian.guendel@fau.de}
\author[1]{Arnaud A. A. Setio}
\author[2]{Florin C. Ghesu}
\author[2]{Sasa Grbic}
\author[2]{Bogdan Georgescu}
\author[3]{Andreas Maier}
\author[2]{Dorin Comaniciu}

\address[1]{Digital Technology and Inovation, Siemens Healthineers, 91052 Erlangen, Germany}
\address[2]{Digital Technology and Inovation, Siemens Healthineers, Princeton, NJ 08540, USA}
\address[3]{Pattern Recognition Lab, Friedrich-Alexander-Universit\"at Erlangen-N\"urnberg, 91058 Erlangen, Germany}


\begin{abstract}
Chest radiography is the most common radiographic examination performed in daily clinical practice for the detection of various heart and lung abnormalities. The large amount of data to be read and reported, with more than 100 studies per day for a single radiologist, poses a challenge in consistently maintaining high interpretation accuracy. The introduction of large-scale public datasets has led to a series of novel systems for automated abnormality classification. However, the labels of these datasets were obtained using natural language processed medical reports, yielding a large degree of label noise that can impact the performance. In this study, we propose novel training strategies that handle label noise from such suboptimal data. Prior label probabilities were measured on a subset of training data re-read by 4 board-certified radiologists and were used during training to increase the robustness of the training model to the label noise. Furthermore, we exploit the high comorbidity of abnormalities observed in chest radiography and incorporate this information to further reduce the impact of label noise. Additionally, anatomical knowledge is incorporated by training the system to predict lung and heart segmentation, as well as spatial knowledge labels. To deal with multiple datasets and images derived from various scanners that apply different post-processing techniques, we introduce a novel image normalization strategy. Experiments were performed on an extensive collection of 297,541 chest radiographs from 86,876 patients, leading to a state-of-the-art performance level for 17 abnormalities from 2 datasets. With an average AUC score of 0.880 across all abnormalities, our proposed training strategies can be used to significantly improve performance scores.
\end{abstract}


\end{frontmatter}

\begin{figure*}

\begin{center}
\includegraphics[width=\linewidth]{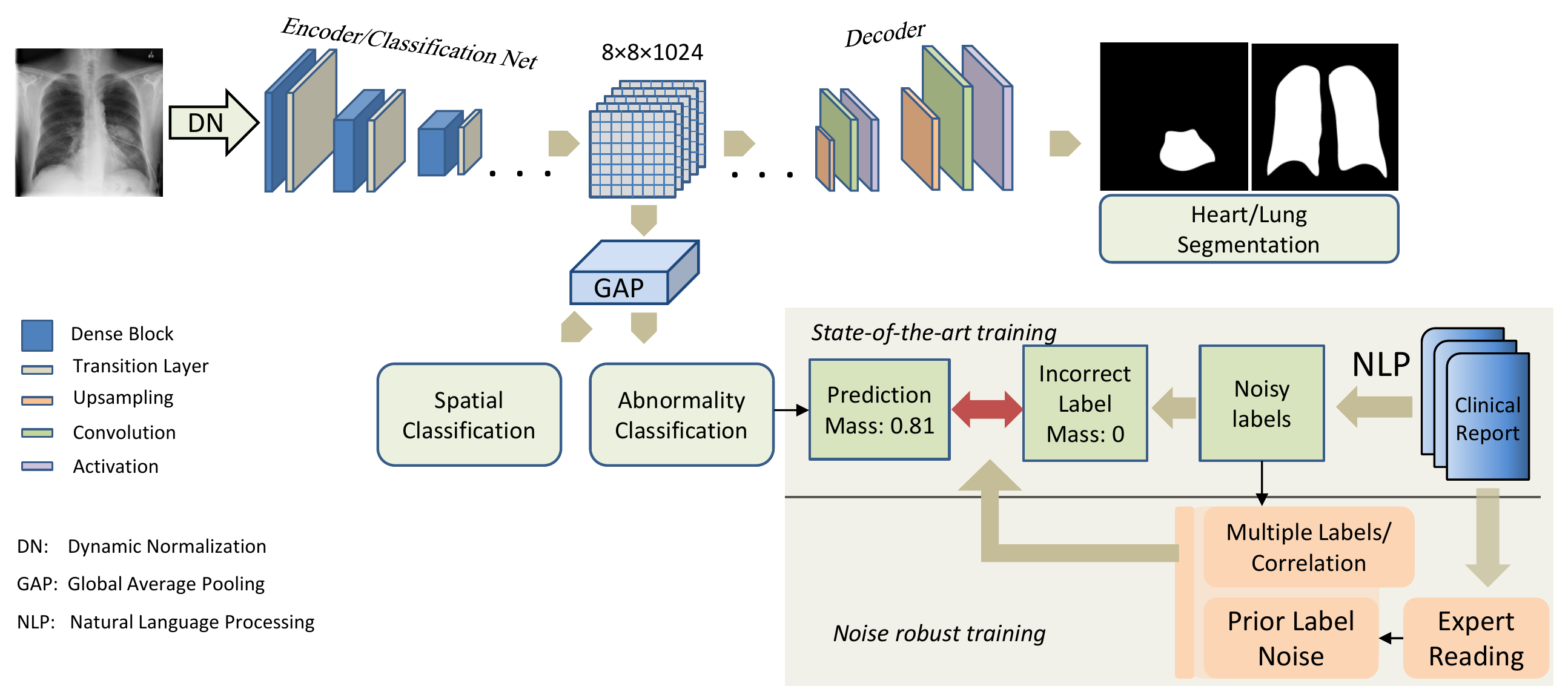}

\ifrevreference
{\hspace{9cm} \color{blue} \rule[3mm]{9cm}{0.4mm} } 
\fi
\caption{Additional Knowledge Integration: We integrate several additional features to handle label noise during training and to increase the abnormality classification performance. \ifrevreference\textbf{C2.4: }\fi\changemarker{Our objective is adapted with prior label noise and label correlation information. In addition to the classification of the abnormalities, spatial classification and lung/heart segmentation are trained.}}
\label{fig:intro}

\end{center}
\end{figure*}

\section{Introduction}
\label{sec:intro}

Recent developments in the deep learning community combined with the availability of large labeled datasets have enabled the training of automated systems that can exceed human performance on a variety of classification, detection and segmentation tasks \citep{Li2016PulmonaryNC, 8187667,Zhu2018DeepLungD3}. In different scenarios, such systems actively support humans, increasing the efficiency and accuracy of their workflow. In the medical domain, deep learning systems for image and data analysis and integration can potentially have an even greater impact, supporting the clinical workflow from patient admission to diagnosis, treatment and follow-up investigations \citep{Rajkomar_2018,Ardila2019EndtoendLC}.
In this paper, we focus on the problem of diagnosing multiple abnormalities based on chest radiography of the human body. In practice, this is a challenging problem reflected in a significant variability between different radiologists \citep{doi:10.1148/rg.2015150023}. For example, a study on determining a diagnosis of tuberculosis between 25 radiologists on 50 chest radiographs leads to a moderate agreement with a kappa score of 0.448 \citep{agreement_study}. The most important cause for high reader variability is the lack of an established guideline of reading (like the Fleischner guideline for CT). As such the interpretation on radiographs is often subjective leading to a large degree of variability. Other factors include the complex appearance of pathologies in radiographs and the large number of scans that need to be read and analyzed daily under time pressure \citep{article_error}. The average time to read and report a plain film is as little as 1.4 minutes \citep{FLEISHON2006453}. 

Aiming to reduce the user variability,  algorithms for automatic classification of abnormalities visible in chest radiographs have been proposed. The recent introduction of large-scale public datasets \citep{wang2017chestx, gohagan2000prostate} has led to series of automated solutions using deep learning for abnormality classification. However, training on these datasets can be suboptimal, as labels were obtained using natural language processed medical reports, which may introduce incorrect labels. As deep learning algorithms tend to overfit to provided (incorrect) labels, networks may be trained into the wrong direction, limiting the performance of the network on unseen data. Moreover, evaluating the performance on a test set with NLP-based labels can be misleading and may not be adequate in representing the performance needed for diagnostic interpretations in practice.

\ifrevreference\textbf{C1.1: }\fi\changemarker{Furthermore, clinical reading reports typically include further relevant information in addition to a found abnormality, e.g., its approximate position. Certain abnormalities, e.g., nodules, appear as small objects in the radiograph.  Thus, prior spatial knowledge about these abnormalities in shape of classification labels (e.g. nodule in the lower-left lung, yes or no?) may regularize the learning process when inductively transferring the information on a shared representation. The resulting learning process, the so-called multi-task learning procedure has been demonstrated in several clinical applications in recent years, e.g. in \citet{multitask}, which leads to improved results on the classification of abnormalities.}

In this study, we propose novel training strategies to handle label noise. First, we evaluate the performance of the NLP labels by performing an observer study, where a subset of training data was re-read by 4 board-certified radiologists. \newline
Thereafter, a novel objective function that takes into account label noise and label correlation is proposed. Label noise was measured from our observer study and label correlation is incorporated to exploit possible comorbidity between abnormalities. 

\changemarker{To regularize the network further, additional tasks are included. Anatomical knowledge is incorporated by adding lung and heart segmentation into the system. Moreover, we also ask the network to not only predict the abnormality, but also the location of the abnormality within the lung. We refer this task as spatial classification.}To deal with multiple datasets and images derived from various scanners applying different post-processing techniques, a novel image normalization strategy is introduced.
 
 The performance of the proposed algorithms was evaluated on a dataset with radiologists-based reference standard. For comparison purposes, performance scores on datasets with NLP-based reference standard is provided. The global architecture and the proposed methods can be seen in Figure \ref{fig:intro}.
 
 \medskip

\textbf{The contributions of this paper are as follows:}
\begin{itemize}
\item{We evaluate and measure label noise by comparing the original NLP-derived labels with the consensus labels from expert readers.}
\item{We formulate a novel objective function that takes into account label noise and label correlation}
\item{We propose a novel multi-task deep neural network for multi-abnormality classification, spatial classification and lung/heart segmentation based on frontal chest radiography images.}
\item{We propose a novel image normalization strategy that robustly handles brightness and contrast variability of images from multiple datasets and scanners.}
\item{We demonstrate that by using the proposed training strategies, additional anatomical knowledge, and novel normalization technique one can significantly increase the accuracy of the abnormality classification.}

\end{itemize}

\section{Related Work}\label{sec:related}
\subsection{Multi-Abnormality Classification}
The publication of the ChestX-ray14 (NIH) dataset  \citep{wang2017chestx}  has led to a series of recent publications that propose automatic systems for abnormality classification. At first,  \citet{wang2017chestx} evaluated several state-of-the-art convolutional neural network architectures, reporting an area under the ROC curve (AUC) of 0.75 on average. \citet{DBLP:journals/corr/IslamAMA17} defined an ensemble of multiple state-of-the-art network architectures to increase the classification performance. Subsequently, \citet{rajpurkar2017chexnet} demonstrated that a common DenseNet architecture \citep{huang2018archive} can surpass the accuracy of radiologists in detecting pneumonia. In addition to a DenseNet, \citet{yao2017learning} implemented a Long-short Term Memory (LSTM) model to exploit dependencies between the abnormalities. An attention guided convolutional neural network architecture was used by \citet{2018arXiv180109927G} to specifically focus on the region of interest which is provided in a second network branch as a cropped image with higher resolution. While \citet{DBLP:journals/corr/abs-1804-07839} designed a Dual-Network to extract the image information of both frontal and lateral views, \citet{DBLP:journals/corr/abs-1807-06067} used a DenseNet architecture integrated with "Squeeze-and-Excitation'' blocks \citep{DBLP:journals/corr/abs-1709-01507} to improve the performance. Moreover, \citet{attention_mining} used a multi-scale aggregation and, additionally, they implemented an attention mining strategy to find accurately diseased regions. \citet{10.1007/978-3-030-00919-9_29} integrated curriculum learning to specifically train based on severity-level attributes of the radiology reports. Attention guidance with heatmaps supported the learning process to increase the performance. \citet{10.1007/978-3-030-00919-9_45} modified the DenseNet architecture to introduce dynamic routing between capsules. Furthermore, \citet{GUAN2018} developed a category-wise residual attention framework with a feature embedding and an attention learning module.
\citet{ChestNet} developed a new network architecture with both a classification and an attention branch, where the latter calculates activation maps with gradient-weighted class activation mapping \citep{DBLP:journals/corr/BaMK14} which is subsequently concatenated with the classification branch. \ifrevreference\textbf{C1.10: }\fi\changemarker{The same group developed a system for channel-wise, element-wise, and scale-wise attention learning to classify 14 abnormalities \citep{TripleAttention}.}Based on very limited location labels of the abnormalities, \citet{DBLP:journals/corr/abs-1711-06373} trained a neural network to predict both classification and localization of the abnormalities. \citet{liu2018sdfn} designed a network architecture with two branches, similar to \citet{2018arXiv180109927G}, where the second branch used a cropped image input based on existing lung masks. \citet{DBLP:journals/corr/abs-1803-07703} defined a network architecture which can be trained on different resolutions. 

In contrast, \citet{10.1371/journal.pmed.1002686} used multiple radiologists to re-read the images: One subgroup of radiologists defined the ground truth of a set where the other subgroup and the neural network was evaluated on. In this way, performance of both the radiologists and the deep learning algorithm could be compared. \citet{irvin2019chexpert} trained on a dataset where the ground truth consists of an additional uncertainty class. Different approaches were applied during training to increase the performance with the uncertainty information. In \citet{fusion_net}, an ensemble of several state-of-the-art architectures is introduced. Additionally, the network is composed of more sub-networks with different resolutions. Finally, \citet{10.1007/978-3-030-32226-7_75} \changemarker{and \citet{GHESU2021101855}} proposed a model that can implicitly learn to estimate predictive uncertainty as an orthogonal measure to the predicted abnormality probability using principles of subjective logic \citep{10.5555/3031657}. \newline
\changemarker{In an attempt to reduce label uncertainty and increase accuracy, the same group proposed an innovative framework for segmentation of airspace opacities from chest radiographs, trained on digitally reconstructed radiographs with precise annotations derived from computed tomography (CT) scans~\citep{barbosa}}\newline
\ifrevreference\textbf{C2.1: }\fi\changemarker{In general, we emphasize that most of the published work report classification results by splitting the data completely randomly for training, validation and testing \citep{yao2017learning,2018arXiv180109927G,DBLP:journals/corr/abs-1711-06373,DBLP:journals/corr/IslamAMA17}. With this splitting strategy, images from the same patient may be located in both training and testing set. For a fair performance evaluation, the splitting should always be performed at patient level. Moreover, to ensure a fair comparison between different methods, the same data split needs to be applied.}

\subsection{Lung and Heart Segmentation}
\ifrevreference\textbf{C1.2/C4.3: }\fi\changemarker{The segmentation of lungs and heart based on chest radiographs has been of high interest in the last years. \citet{multi-organ-seg} proposed a method to segment the lungs and the heart including a novel network architecture based on a U-net as backbone. Another method detects the lung boundaries with training features of the gray level co-occurrence matrix \citep{lung_boundary_det}. \citet{lung_seg_critic} developed an architecture for lung segmentation using an adversarial critic network, in addition to the segmentation network. In the work of \citet{contour_seg}, different state-of-the-art segmentation networks are used as backbone. In this method, the contours of the lungs were determined and used for training. \citet{dense_segment} specifically focused on chest radiographs with high intensity change because of the abnormalities resulting in extensively dense regions. \citet{9126830} demonstrated that an additional variational autoencoder fulfilled anatomical plausibility which improved the segmentation of lungs and heart.} \newline

\subsection{Spatial Classification}
\changemarker{For spatial classification there is currently no research available as the chest radiography datasets typically include either no additional spatial information or precise detection information which can be converted into segmentation maps.}\\

\subsection{Noisy Label Classification}
\changemarker{While many publications highlight the high fraction of label noise in the ChestX-ray14 dataset, few solutions have been proposed to address this limitation. \citet{label_noise_robust} showed that, to some extent, deep learning models can be robust against label noise during training, in general. Therefore, many groups kept standard training procedures and focused solely on the re-definition of labels in the test set \citep{10.1371/journal.pmed.1002686}. An overview of possibilities to deal with label noise on medical image data was demonstrated by \citet{label_noise_overview}. The methods for noise robustness were defined in different categories, e.g., network architecture and training procedures. In terms of loss adaption, the categorical cross-entropy was changed by \citet{cce_noise}, adding what they called the least trimmed absolute value estimator which makes the training more robust to outliers. Most publications in the field of label noise robustness, however, show their approaches based on standard datasets \citep{10.5555/3305890.3305939,10.5555/3295222.3295387,pmlr-v48-gal16}, e.g., MNIST and \mbox{CIFAR}, remaining unclear how well these methods can generalize on medical datasets.}\\

\section{Problem Definition and Dataset Analysis}
\label{sec:Classification}

In this Section, the datasets that are used to train and validate our method are presented. The observer study aimed to evaluate the label noise on these datasets is described. Thereafter, correlation between labeled abnormalities is evaluated.

\subsection{Dataset}
\label{subsec:dataset}

\begin{figure*}[t]
\begin{center}
\includegraphics[width=4in]{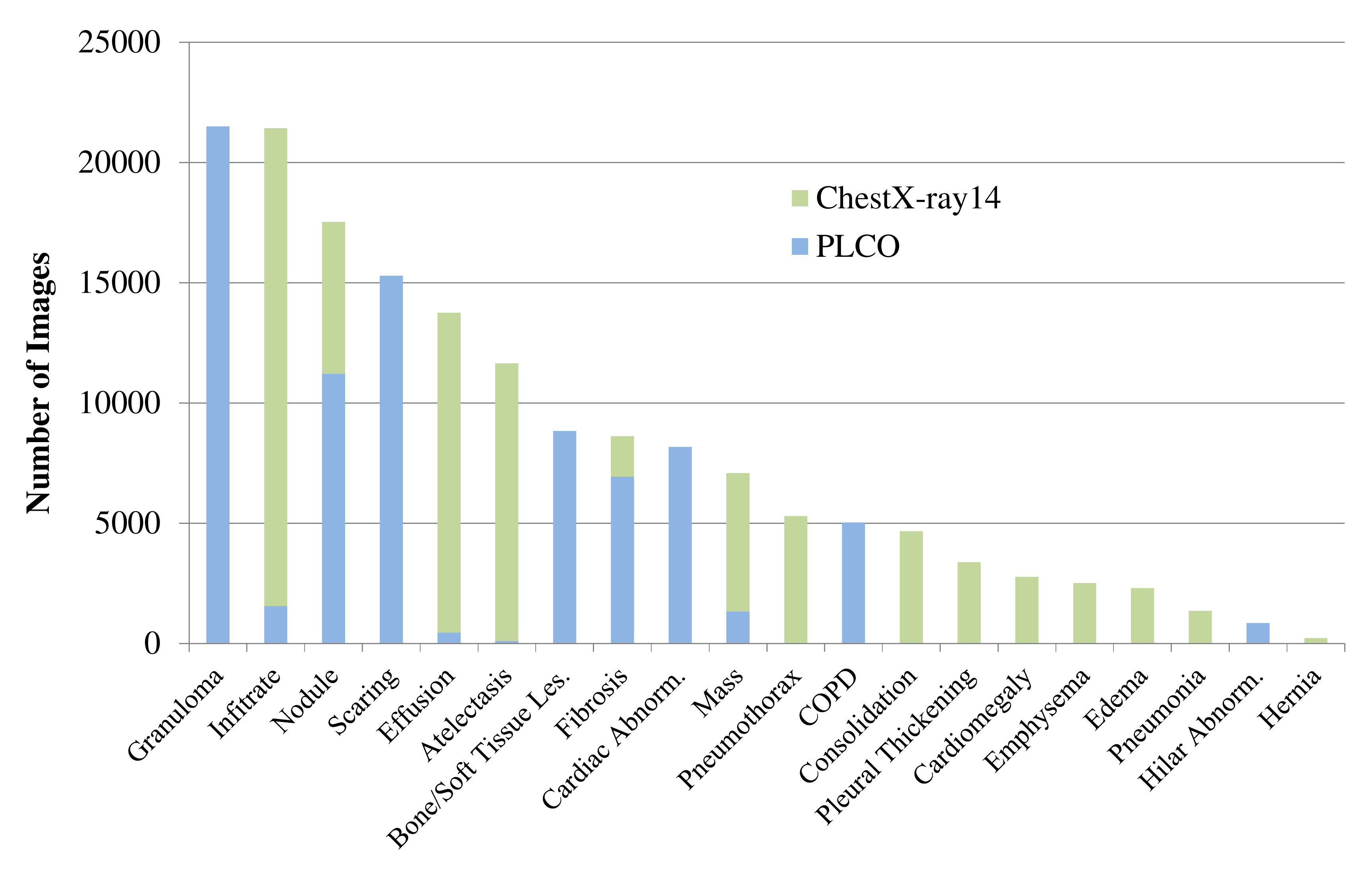}
\caption{The number of images associated with abnormalities in ChestX-ray14 and PLCO datasets. The number of images where none of these pathologies appear is excluded.}
\label{fig:num_images}
\end{center}
\end{figure*}

Our data collection is composed of two different datasets, the ChestX-ray14 \citep{wang2017chestx} and PLCO \citep{gohagan2000prostate} dataset. The ChestX-ray14 database contains 67,310 posterior-anterior (PA) and 44,810 anterior-posterior (AP) images whereas the PLCO dataset is based on a screening study and solely contains PA images. Table \ref{tab:dataset_overview} gives an overview of the datasets. By combining both datasets, 297,541 frontal chest radiographs from 86,876 patients can be used. Including follow-up scans, there is an average of 3-4 images per patient. The PLCO dataset includes spatial information, i.e. approximate knowledge about where abnormalities are located. Figure \ref{fig:num_images} shows the number of images that contain each abnormality. One image can also show multiple abnormalities. Additionally, the collection contains 178,319 images where none of the mentioned abnormalities appear, these images are not counted in Figure \ref{fig:num_images}. Note that the high imbalance of the data collection with respect to different abnormalities represents a challenge in ensuring training stability and performance.

\begin{table*}[!ht]
\small
\caption{\changemarker{Overview of the 2 considered datasets. The combined data collection consists of 297,541 images from 86,876 patients. The last column describes the subset which is read by our expert radiologists. Note that the image number (first row) denotes the entire number of images in the dataset, however, patients used for the expert read are excluded in the ChestX-ray14 dataset (first column) for training and validation. The last two rows show the split statistics for training (train), validation (valid), calibration (calib), and testing (test).}}
\begin{center}

\begin{tabu}{L{3.3cm} | C{1.4cm} C{1.4cm} | C{1.4cm} C{1.4cm} C{1.4cm} | C{1.4cm} C{1.4cm}} 

Name & \multicolumn{2}{c|}{\textbf{ChestX-ray14}} & \multicolumn{3}{c|}{\textbf{PLCO}} & \multicolumn{2}{c}{ \begin{tabular}{@{}c@{}}\textbf{Consensus Expert Read} \\ \textbf{(ChestX-ray14)}\end{tabular}}  \\
Image / Patient number (\#) & \multicolumn{2}{c|}{112,120 / 30,805} & \multicolumn{3}{c|}{185,421 / 56,071} & \multicolumn{2}{c}{689 / 689}  \\
Abnormality number & \multicolumn{2}{c|}{14} & \multicolumn{3}{c|}{12} & \multicolumn{2}{c}{5}  \\
Image size & \multicolumn{2}{c|}{$1024\times1024$} & \multicolumn{3}{c|}{$\sim2500\times2100$} & \multicolumn{2}{c}{$1024\times1024$}  \\
\hline
\rowfont{\color{black}}Split & train & valid & train & valid & test & calib & test \\

\rowfont{\color{black}}Image number (\%) & 90 & 10 & 70 & 10 & 20 & 30 & 70\\
\rowfont{\color{black}}Image number (\#) & 99,462 & 8,338 & 129,658 & 18,773 & 36,990 & 207 & 482 \\

\end{tabu}
\\
\end{center}

\label{tab:dataset_overview}
\end{table*}

\ifrevreference\textbf{C2.1: }\fi\changemarker{\textbf{Assessing the Quality of Labels:}}As the ChestX-Ray14 dataset contains 112,120 images, re-reading entire dataset is not feasible. Therefore, we selected a subset of samples using random sampling. We limit the process to 5 abnormalities (effusion, cardiomegaly, consolidation, atelectasis, mass) which were chosen based on clinical importance. We take only one image per patient. The expert reading subset contains 689 images. The remaining images of each patient are removed from the remaining data to avoid a patient overlap. The reading process was based on consensus decision-making: First, 4 board-certified radiologists blindly re-labeled the images. In a second stage, for all cases where consensus was not reached on all labels through the independent read, an open discussion was carried out to establish consensus labels. The original dataset labels were not provided during the re-reading process to avoid a biased decision towards the original labels.

We hypothesize that the re-reading process following of consensus of experts is imperative to understand and address the limited quality of the labels of the ChestX-Ray14 dataset. Thus, we assume that

\begin{equation}\label{eq:corr2}
\begin{split}
Y^{True} \approx Y^{Rad},
\end{split}                    
\end{equation}

where $Y^{True}$ are the true (unknown) labels, and $Y^{Rad}$ the generated labels by consensus of expert radiologists. An overview of the subset can be seen in the last column of Table \ref{tab:dataset_overview}.

\textbf{Dataset Splits:} As Table \ref{tab:dataset_overview} shows, 2 datasets are used and a subset of \ifrevreference\textbf{C1.4:}\fi\changemarker{ChestX-ray14 images are re-read to obtain higher quality reference labels. }The PLCO dataset is split into 70\% for training, 10\% for validation, and 20\% for testing. The ChestX-ray14 dataset (excluding the re-read samples) is split into 90\% training and 10\% validation. A subset of the ChestX-ray14 set is re-read (see last column in Table \ref{tab:dataset_overview}) where 70\% is used for the evaluation. The last 30\% serve as calibration of our proposed methods. \textit{Patient-wise} splits are considered for all experiments to separate the patients into training, validation, and test set.

\subsection{Errors in the Abnormality Labeling Process}
\label{subsec:error_domain}

The misinterpretation of medical images can be caused by different reasons, e.g, radiologist fatigue or lack of attention \citep{article_error}. Recently, many publications highlight the significant label error in the ChestX-ray14 dataset \citep{ChestXray14Problems}. Beside the clinical reader, another major error domain is derived from the natural language processing (NLP) algorithm that transfers the clinical reports into binary abnormality labels. 

\begin{figure}[t]
\begin{center}
\includegraphics[width=\linewidth]{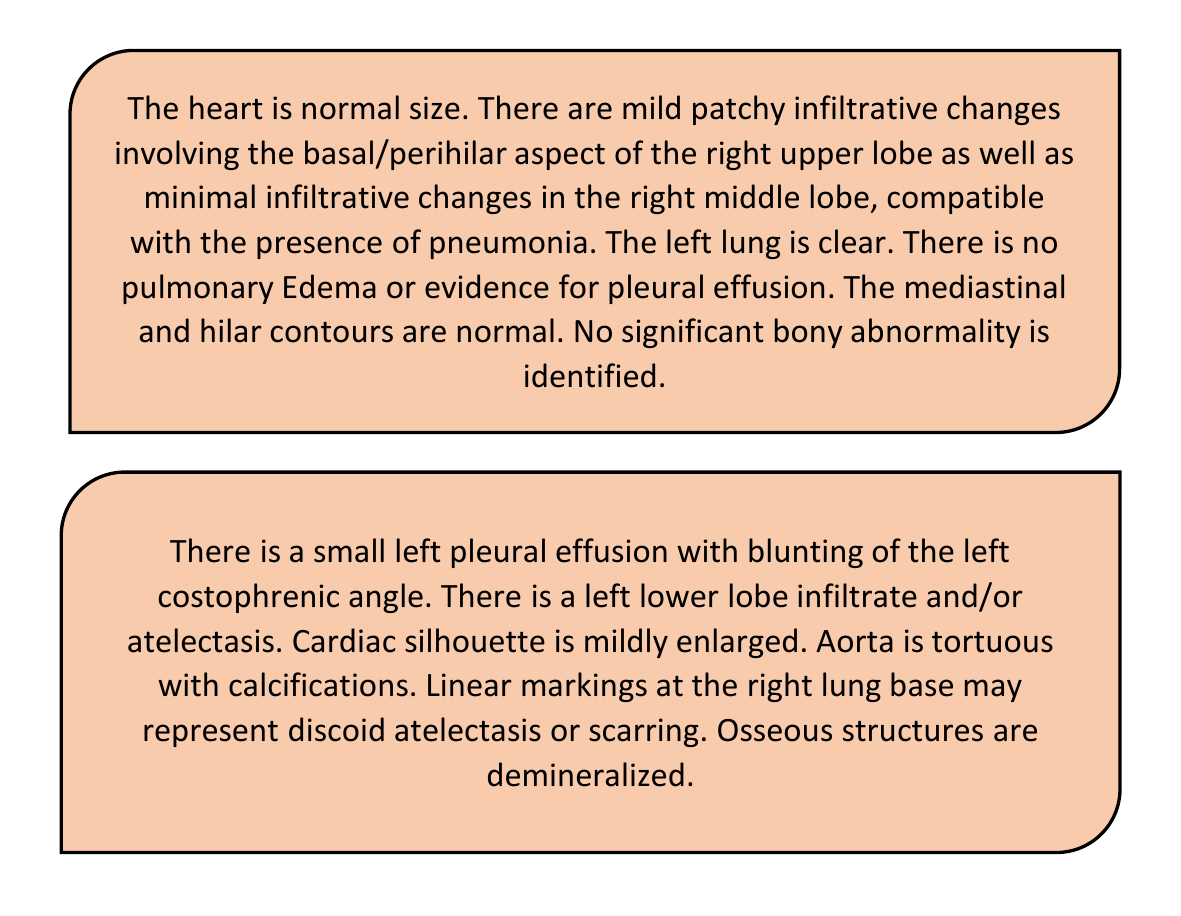}
\caption{Two randomly chosen, abnormal clinical reports based on chest radiographs derived from the same reader.}
\label{fig:reports}
\end{center}
\end{figure}

Based on two reports from the same reader in Figure \ref{fig:reports}, the complexity and variability of the content can be observed. As medical reports for chest radiography do not follow well-established guidelines, the structure of the content is typically built up in arbitrary order. Furthermore, we see ambiguous terms, e.g., \textit{``infiltrate and/or atelectasis''} or \textit{``may represent''}. We hypothesize that this ambiguity combined with the unstructured complexity and additional user variance may lead to errors in the natural language processing algorithm and, thus, the noise in the dataset labels \citep{NLPamb}.

This error leads to an incorrect reference standard on which the networks are trained and evaluated. Several publications show that this label error negatively affects the performance, e.g., in classifying abnormalities correctly \citep{10.1371/journal.pmed.1002686}. However, the degree of noise that is present in these datasets remains unclear. In order to measure the degree of label noise in our datasets, we perform an observer study on a subset of data and quantitatively assess the label noise probabilities.

\begin{figure*}
\begin{center}
\includegraphics[width=\linewidth]{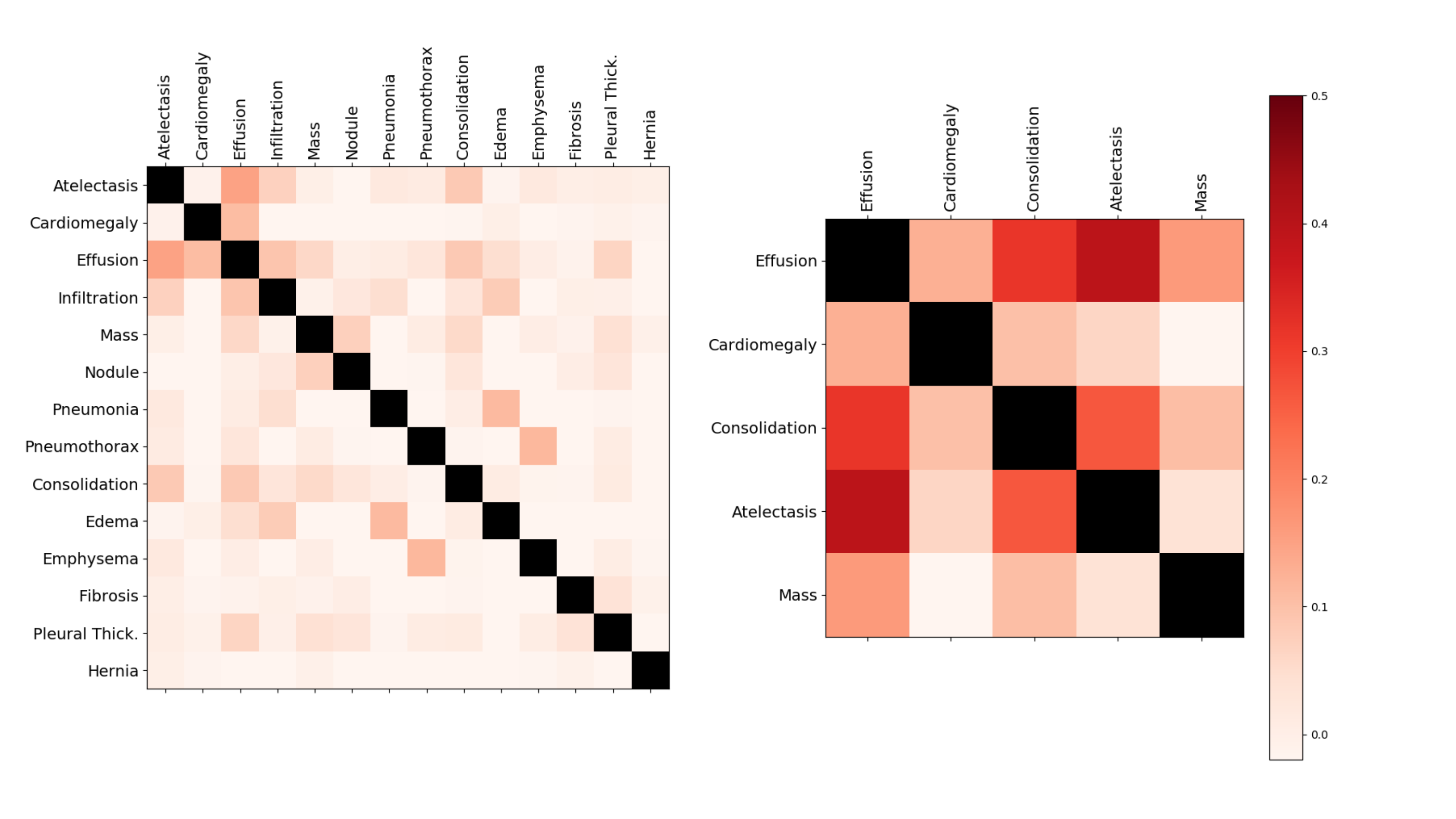}
\caption{Pearson correlation of abnormalities; Left: Correlation of the original ChestX-Ray14 dataset labels; Right: Correlation of redefined labels based on a consensus of radiologists.}
\vspace{-0.15in}
\label{fig:corr}
\end{center}
\end{figure*}

\subsection{Exploiting Abnormality Comorbidity: A Correlation Perspective}
\label{subsec:joint_class}

The comorbidity of abnormalities have been explored in many fields, e.g., in \citet{Guan2000547}. Comorbidity indicates simultaneous presence of multiple abnormalities. The knowledge of comorbidity may assist clinicians in making a more precise assessment of diseases or abnormalities \citep{comorbid}. In this section we present a principled strategy of how to exploit the comorbidity of different abnormalities to improve the robustness of the model.

In the context of multi-label classification, the correlation of different labels is an important piece of information \citep{Ghamrawi:2005:CMC:1099554.1099591, zhu}. Conceptually, we analyse how strong a set of class labels  $c^{(n)}$ for abnormality $n$ correlate with a set of class labels $c^{(r)}$ for abnormality $r$  where $r \in \{1 \dots D\} \setminus \{n\}$ and $D$ denotes the number of abnormalities. We use the Pearson correlation coefficient which is sensitive to class imbalance to measure the correlation on the ChestX-Ray14 training set between the abnormalities:

\begin{equation}\label{eq:corr1}
\begin{split}
corr_{Pearson}\left(c^{(n)},c^{(r)}\right) = \dfrac{cov\left(c^{(n)},c^{(r)}\right)}{\sigma^{(n)}\sigma^{(r)}},
\end{split}                    
\end{equation}

where $cov$ denotes the covariance and $\sigma$ the standard deviation. In Figure \ref{fig:corr} (left), we color the strength of correlation between the abnormalities of the original dataset labels. The correlation of the dataset with labels based on a consensus of radiologists can be seen in Figure \ref{fig:corr} (right) which shows stronger correlation between the abnormalities, e.g., the Pearson correlation coefficient between atelectasis and effusion equals 0.395. 

There are different reasons why the dataset labels have significant correlation: Based on the paper by \citet{irvin2019chexpert}, abnormalities can be represented hierarchically. Thus, an abnormality can consist of a subgroup of abnormalities, e.g., lung opacity includes pneumonia or edema. This stratification into different groups is also addressed in \citet{ChestXray14Problems2}. Further existing publications analyse the relationships between abnormalities, e.g., lung inflammation and cardiovascular disease \citep{doi:10.1164/rccm.201203-0455PP}.

Overall, we hypothesize that the given correlation between abnormalities may be related to some degree of existing comorbidity. In practice, this correlation information can be exploited to serve an an additional learning support during training.

\section{Methodology}
\label{sec:joint}

Given an arbitrary AP or PA chest radiograph $\mathbf{I}$ with size $N\times N$ pixels, we design a deep learning based system parametrized by $\theta$ which outputs the probability of different abnormalities being present in the image: $\vec{o} = p(\mathbf{I}; \theta)$, where $\vec{o}\in[0,1]^D$ and $D$ is the number of considered abnormalities. 

We exploit the correlation between abnormalities and regularize our loss function using information from both label noise and label correlation. The system is also designed to predict approximate spatial location of the abnormalities and a probabilistic segmentation map $\mathbf{S}\in[0,1]^{2\times N\times N}$ for both lungs and the heart. The proposed normalization strategy wraps up our novel multi-task system focusing on the improvement of abnormality classification. Figure \ref{fig:intro} shows the global architecture including our proposed methods.

\subsection{Deep Neural Network Design}
\label{subsec:NetworkDesign}
The backbone architecture of our model is inspired from the DenseNet architecture \citep{huang2018archive}. We adopt this network architecture with 5 dense blocks and a total of 121 convolutional layers. Each dense block consist of several dense layers which include batch normalization, rectified linear units, and convolution. The novelty of the DenseNet are the skip connections, meaning that within a block, each layer is connected to all subsequent layers. Between each dense block, a so-called transition layer is added, which includes batch normalization, convolution, and pooling, to reduce the dimensions. \par 

The single grayscale input image $\mathbf{I}$ is rescaled to  $N\times N$ pixels (in our experiments $N$=256) using bilinear interpolation and fed into the network. \ifrevreference\textbf{C2.3: }\fi\changemarker{The network is initialized with the pre-trained ImageNet model \citep{ImageNet}. To maintain all pre-trained weights, we replicate the image to 3 channels.} The global average pooling (GAP) layer is applied to obtain the representation to be used for classification. The number of output units is set to the number of abnormality classes $D$. We use sigmoid activation functions for each class to map the output to a probability interval $[0,1]$.

\subsection{Model Training}
\label{subsec:Training}
The algorithm for multi-label classification problem is trained using the following approach: The training process is modified such that each class can be trained individually. We create \textit{D} binary cross-entropy loss functions. The corresponding labels $[c^{(1)}, c^{(2)} \ldots c^{(D)}] \in \{0,1\}$ (absence or presence of the abnormality, respectively) are compared with the network output $[p^{(1)}, p^{(2)} \ldots p^{(D)}] \in[0,1]$ and the loss is measured. Due to the highly imbalanced problem we introduce additional weight constants $w_P^{(n)}$ and $w_N^{(n)}$ for each abnormality indexed by \textit{n}, to the cross-entropy function:\\

\begin{equation}\label{eq:loss1_g}
\begin{split}
\mathcal{L}_{Abn} = -\sum_{n=1}^{D} \sum_{i=1}^{F} \Big[ &w_P^{(n)} c^{(n)}_i\ln\left(p^{(n)}_i\right) +  \\
&w_N^{(n)} \left(1 - c^{(n)}_i\right)\ln\left(1 - p^{(n)}_i\right) \Big],
\end{split}                    
\end{equation}

where $w_P^{(n)} = \frac{P^{(n)} + N^{(n)}}{P^{(n)}}$ and $w_N^{(n)} = \frac{P^{(n)} + N^{(n)}}{N^{(n)}}$, with $P^{(n)}$ and $N^{(n)}$ indicating the number of positive and negative cases for the entire training dataset, respectively. \ifrevreference\textbf{C1.5: }\fi\changemarker{The same weighting strategy is also applied in \citet{wang2017chestx} and \citet{10.1371/journal.pmed.1002686}.}Equation \ref{eq:loss1_g} calculates the loss sum over all images indexed by $i$, where $F$ denotes the total number of images in the set. For all experiments, we train with 128 samples in each batch. The Adam optimizer \citep{adam} ($\beta_1 = 0.9$, $\beta_2 = 0.999$, $\epsilon = 10^{-8}$)  is used with an adaptive learning rate: the learning rate is initialized with $10^{-3}$ and reduced by a factor of 10 when the validation loss plateaus. All parameters can be found in Table \ref{tab:hyperparameters}.\\

\begin{table}[t]
\small
\caption{Hyperparameters used to train our model.}
\label{tab:hyperparameters}
\begin{center}
\begin{tabular}{ l  l } 
\hline
Hyperparamter & Value \\
\hline
Initial Learning Rate & $10^{-3}$ \\

Image Input Size & 256\\

Batch Size & 128 \\

Optimizer &  Adam \\

Loss Function & Binary Cross Entropy\\

Number of Epochs & 30 (early stopping) \\
\hline

\end{tabular}
\\
\end{center}

\vspace{-0.15in}
\end{table}

An important issue encountered with abnormality labeling is the varying and overlapping definition and interpretation between radiologists as seen in Section \ref{subsec:error_domain}. Therefore, we treat the corresponding abnormalities of both datasets separately for the experiments in Section \ref{sec:Experiments} and create different output classes. Given $D_1=14$ abnormalities of the ChestX-ray14 dataset and $D_2=12$ abnormalities of the PLCO dataset, we define $D=D_1+D_2=26$ classes for our network. Furthermore, we only compute gradients for labels of one dataset where the current image is derived from. This strategy avoids a class categorization step beforehand and ensures that each network layer (except the last) receives information of all images.

\subsection{Prior Label Noise Probability}

\begin{table}
\small
\caption{Sensitivity/Specificity scores of original dataset labels versus re-read labels.}
\begin{center}
\begin{tabular}{ l  c  c} 

\hline

Abnormality & $s_{sens}$ & $s_{spec}$\\
\hline

Effusion & 0.300 & 0.966\\
Cardiomegaly & 0.342 & 0.986\\
Consolidation & 0.129 & 0.949\\
Atelectasis & 0.221 & 0.970\\
Mass & 0.364 & 0.972 \\
\hline
Average & 0.271 & 0.969\\
\end{tabular}
\\
\end{center}

\label{tab:sens_spec}
\end{table}

One can compute sensitivity and specificity of the original dataset labels based on the subset of re-read labels. Assuming the corrected labels are the true labels as discussed in Section \ref{subsec:dataset}, we compute the performance scores which reflect the noise probabilities of positives and negatives on the original ChestX-Ray14 labels. \ifrevreference\textbf{C1.6: }\fi\changemarker{Table \ref{tab:sens_spec} shows sensitivity $s_{sens}=\frac{TP}{P}$ and specificity $s_{spec}=\frac{TN}{N}$ scores of the selected abnormalities which are used to regularize the loss function. TP and TN respectively denote the number of original positive and negative labels which are correctly labeled based on the re-read subset. Parameter P and N stand for the total number of positive and negative cases in the re-read subset. Thus,}low scores indicate strong label noise. We adapt our original loss function with a second term which is based on the inverse binary cross entropy function:

\begin{equation}\label{eq:Noise1}
\begin{multlined}
\mathcal{L}_{Noise} = \mathcal{L}_{Abn} + r_{noise}\\
= -\sum_{n=1}^D \sum_{i=1}^F \bigg[ w_P^{(n)} c_i^{(n)} \ln p_i^{(n)} + w_N^{(n)}\left(1-c_i^{(n)}\right)\ln\left(1-p_i^{(n)}\right) +\\
\quad\;\; \lambda_{Noise} \Big[f_P^{(n)}w_N^{(n)}\left(1-c_i^{(n)}\right) \ln p_i^{(n)} +\\ f_N^{(n)}w_P^{(n)}c_i^{(n)} \ln\left(1-p_i^{(n)}\right) \Big]\bigg],
\end{multlined}                    
\end{equation}

where $f_P$ and $f_N$ are the individual regularization weights for positive and negative examples. In our experiments, we set $f_P^{(n)}=1-s_{sens}^n$ and $f_N^{(n)}=1-s_{spec}^n$. The parameters are derived based on the calibration set which is introduced in Section \ref{subsec:dataset}. Parameter $\lambda_{Noise}$ is another weight to define the overall influence of the regularization term. Strong weights mean that a higher error is computed when predicting the original label. In our experiments, we set $\lambda_{Noise} = 0.1$ (value with best performance on validation set). The regularization term is disregarded for abnormalities excluded in the calibration set.

\subsection{Prior Label Correlation}

In Figure \ref{fig:corr}, we see that the labels of many classes are correlated. This correlation information can also be used as another prior in the loss function. We adapt the original loss function with a term to consider the information across all abnormality labels:

\begin{equation}\label{eq:Noise2}
\begin{split}
& \mathcal{L}_{Corr} = \mathcal{L}_{Abn} + r_{corr} \\ 
& = -\sum_{i=1}^F \sum_{n=1}^D \bigg[w_P^{(n)} c_i^{(n)} \ln p_i^{(n)} + w_N^{(n)} \left(1-c_i^{(n)}\right)\ln \left(1-p_i^{(n)}\right) +\\
&  \:\;\sum_{r\in \{1\dots D\}\setminus \{n\}} \!\!\! \Big[ cov^{(n,r)} \Big[w_P^{(r)}c_i^{(r)} \ln p_i^{(r)} +\\ 
& \qquad \qquad \:\: w_N^{(r)} \left(1-c_i^{(r)}\right)\ln\left(1-p_i^{(r)}\right)\Big]\Big]\bigg],
\end{split}                    
\end{equation}

where $cov^{(n,r)}$ with element $(n,r)$ measures the covariance between label indexed as $n$ and the label indexed as $r$. Thus, label correlation between 2 abnormalities influence on the overall loss function. Depending on the covariance matrix, all abnormality labels may influence on a specific abnormality. The covariance matrix is calculated based on the calibration subset. The regularization term is disregarded for abnormalities excluded in the calibration set.

\subsection{Integrating Additional Knowledge}
\label{subsec:additional_knowledge}

Additional \changemarker{anatomical}knowledge related to individual pathologies as well as the underlying anatomy, i.e., the heart and the lungs, can be exploited to increase the classification performance.

\subsubsection{Lung and Heart Segmentation}
\label{subsec:segmentation}
First, one can focus the learning task to the heart and lung regions. The image information outside of these regions may be regarded as irrelevant for the diagnosis of these lung/heart abnormalities. Lung and heart segmentation masks are  available for the entire data collection.
\par

\begin{figure}[t]
\begin{center}
\includegraphics[width=\linewidth]{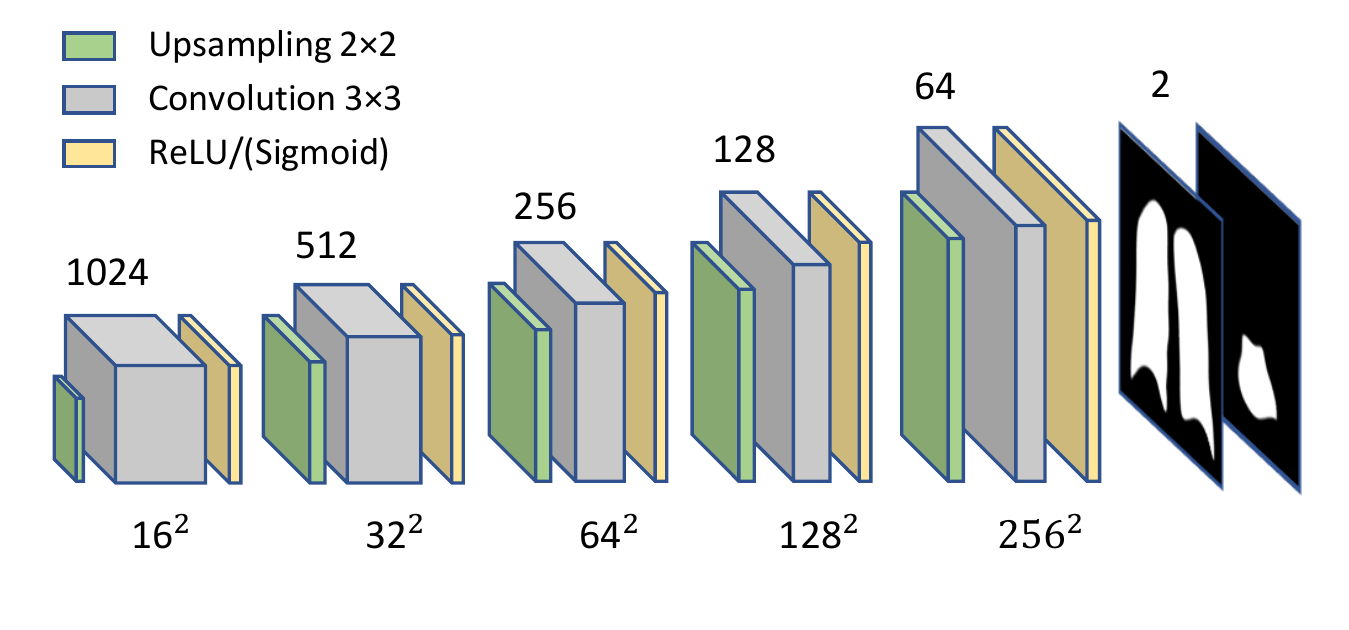}
\textbf{}\caption{Architecture of the decoder to predict segmentation masks. The top number shows the channel number, the bottom value indicates the feature map size. The network is connected to the classification network after the last dense block (left). The final sigmoid layer predicts the lung and heart masks in 2 channels (right).}
\label{fig:decoder}
\end{center}
\end{figure}

Instead of providing the masks as input for the classification network, we extend the classification network with a decoder branch and predict the masks (see Figure \ref{fig:intro}).
In this way, the additional knowledge about the shape of the heart and lungs is integrated in an implicit way, i.e., during learning through the flow of gradients. As such, in the encoder part, the network learns features that are not only relevant for the abnormality classification, but also for the isolation/segmentation of the relevant image regions.

\par

The DenseNet model described in subsection \ref{subsec:NetworkDesign} is extended to solve the segmentation task. Therefore, we add a decoder network whose input is the returning feature maps of the last dense block (see Figure \ref{fig:intro}). The decoder architecture is visualized in Figure \ref{fig:decoder}. For the segmentation task, we use the mean squared error loss function:

\begin{equation}\label{eq:mseloss}
\mathcal{L}_{Seg} = \sum\limits_{i=1}^F \left[\frac{1}{t}\sum\limits_{z=1}^t \left(s_{i,z} - p_{i,z}\right)^2\right],
\end{equation}

where $t=2\times N\times N$ and $p_{i,z}\in \mathbf{S}$ denotes the output prediction of the current pixel $z$ and $s_{i,z}\in\{0,1\}$ the corresponding pixel label.

\subsubsection{Spatial \changemarker{Classification}}

We propose to add additional supervision during learning using several approximate spatial labels provided with the PLCO data. For five abnormalities (Nodule, Mass, Infiltrate, Atelectasis, Hilar Abnormality) coarse location information is made available (see Table \ref{tab:region}). 

\begin{table}[t]
\small
\caption{Spatial Class Labels for the PLCO data}
\begin{center}
\begin{tabular}{ c  l c  l } 

\hline
No. & Region & No. & Region\\
\hline

1 & Left lung  & 6 & Upper-middle part \\
2 & Right lung & 7 & Upper part\\
3 & Lower part & 8 & Diffused\\ 
4 & Lower-middle part & 9 & Multiple (more  \\
5 & Middle part & & independent parts)\\

\hline

\end{tabular}
\\
\end{center}

\label{tab:region}
\vspace{-0.15in}
\end{table}

The location information is incorporated in the cross-entropy loss using location-specific classes. The spatial labels $[b_1, b_2 \ldots b_L] \in \{0,1\}$, where \textit{L} is the total number of spatial classes listed in Table \ref{tab:region}, are compared with the network prediction and the loss is calculated (Equation \ref{eq:loss2}).

\begin{equation}\label{eq:loss2}
\begin{split}
\mathcal{L}_{Loc} = -\sum_{m=1}^{L} \sum_{i=1}^F \Big[&w_P^{(m)} b^{(m)}\log\left(p^{(m)}\right) + \\
&w_N^{(m)} \left(1 - b^{(m)}\right)\log\left(1 - p^{(m)}\right)\Big],
\end{split}                    
\end{equation}

where $w_P^{(m)} = \frac{P_m + N_m}{P_m}$ and $w_N^{(m)} = \frac{P_m + N_m}{N_m}$, with $P_m$ and $N_m$ indicating, respectively, the number of presence and absence cases of spatial class \textit{m} in the training set. The individual localization loss
$\mathcal{L}_{Loc}$ is activated/deactivated dynamically: If spatial labels are not available for abnormality $n$, all spatial labels are disregarded and no gradients are computed. Otherwise, the loss is calculated as Equation \ref{eq:loss2} shows. 

\subsection{Dynamic Normalization}
One challenge in processing chest radiographs is accounting for the large variability of the image appearance, depending on the acquisition source, radiation dose as well as proprietary non-linear postprocessing. In practice, one cannot systematically address this variation due to missing meta-information (e.g., unknown maximum high voltage for images from the ChestX-ray14 dataset). In this context, generic solutions have been proposed for the normalization of radiographs using multi-scale contrast enhancement/leveling techniques \citep{Philipsen2015LocalizedEN, 1000258}.

\begin{figure}[t]
\begin{center}
\includegraphics[width=\linewidth]{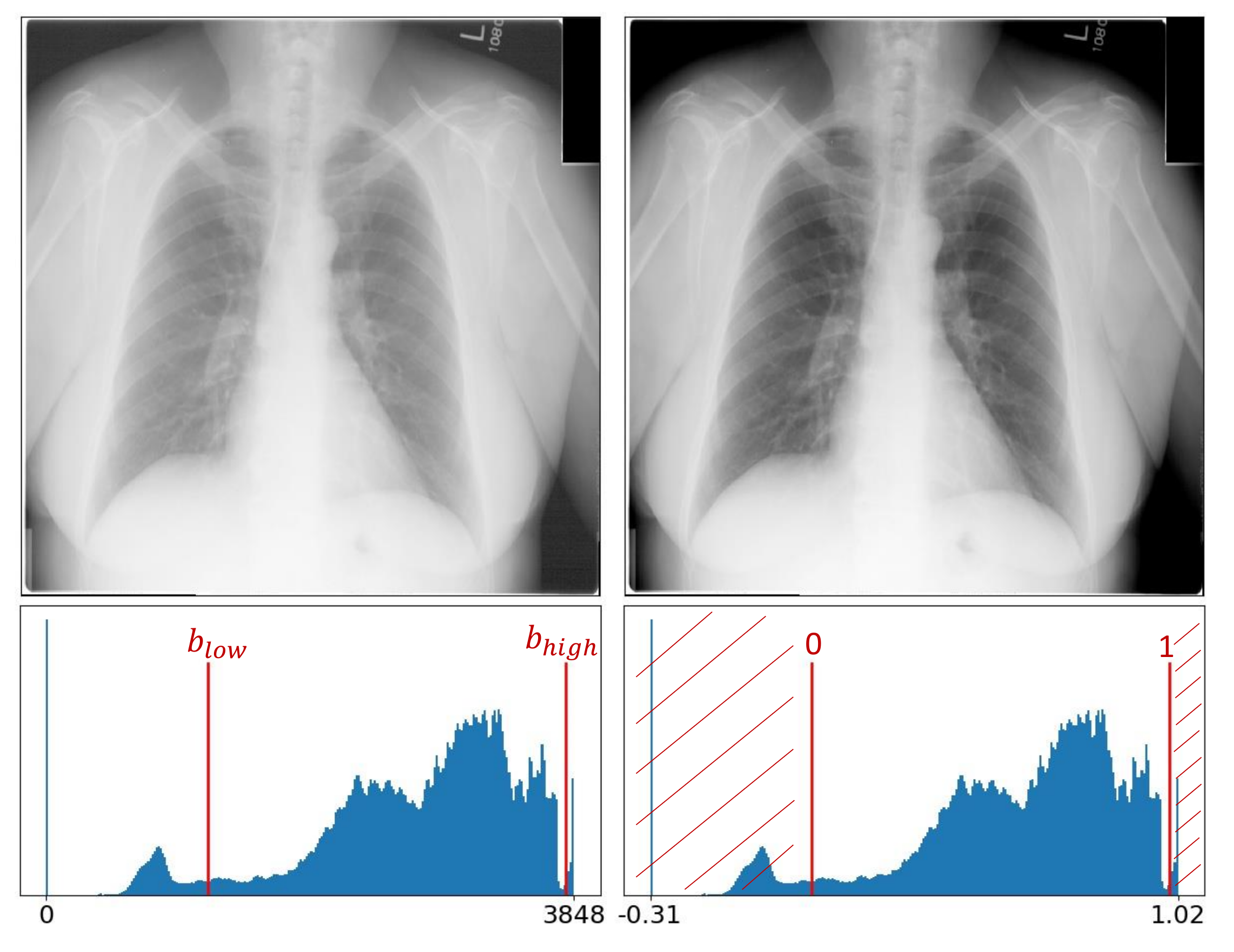}
\caption{An original image of the dataset and the corresponding intensity histogram (\textit{x-axis:} intensity value; \textit{y-axis:} number of pixels) is displayed (left). The preprocessed image where the described normalization technique based on the bounds $b_{low}$ and $b_{high}$ is applied (right). The high signal spike in the left histogram at value 0 represents the black anonymization box.}
\label{fig:norm}
\end{center}
\end{figure}

For our diagnostic application, we propose to explicitly avoid altering the image appearance using one of these methods. Instead, we propose an efficient method for dynamically windowing each image, i.e., adjust the brightness and contrast via a linear transformation of the image intensities. Given an arbitrary chest radiograph $\mathbf{I}$, let us denote its pixel value histogram function as $h(x;\mathbf{I})$. We use a bandwidth of 256 for the image intensity histogram function. Using Gaussian smoothing and median filtering, one can significantly reduce the noise of $h$ (visible as, e.g., signal spikes due to black background or white text overlay) as well as account for long function tails that affect the windowing of the image. As such, based on the processed function $h$, we determine two bounds $b_{low}$, and $b_{high}$ which represent a tight intensity window for image I. The value $b_{low}$ indicates the lowest bin and $b_{high}$ the highest bin along the image intensity histogram. The normalization is applied as follows, $\mathbf{I} = (\mathbf{I} - b_{low}) / (b_{high} - b_{low})$. A visual example is shown in Figure \ref{fig:norm}. 

\section{Experimental Results}
\label{sec:Experiments}

In our experiments, we measured the performance of our baseline system (\textbf{baseline}) presented in Section \ref{subsec:NetworkDesign} and \ref{subsec:Training}, and quantified the improvement achieved by 1) including the image normalization component (\textbf{norm}), 2) segmenting the heart/lung region (\textbf{seg}), 3) approximately localizing pathologies within the image (\textbf{loc}), and 4) adding the regularized loss terms to compete with the noisy labels (\textbf{noise}/\textbf{corr}). All results can be seen in Table \ref{tab:classresults_all}. Finally, we show the performance of 2 joint models: one including normalization, anatomy information and noise regularization (\textbf{baseline} + \textbf{norm} + \textbf{seg} + \textbf{loc} + \textbf{noise}); and the other including normalization, anatomy information and correlation regularization (\textbf{baseline} + \textbf{norm} + \textbf{seg} + \textbf{loc} + \textbf{corr}). We refer to the first joint model as \textbf{all-noise} and the to the second as \textbf{all-corr}. \ifrevreference\textbf{C1.7: }\fi\changemarker{The column \textbf{all} includes all previous features and both regularization schemes.} Please recall that we use 70\% of the consensus expert reading subset (ChestX-ray14) and 30\% of the PLCO dataset. All experiments were measured with the area under the curve (AUC). Corresponding statistical tests \citep{PMID:3203132} were conducted to measure whether the improvement \ifrevreference\textbf{C2.7: }\fi\changemarker{between the baseline and the best model} is significant. \changemarker{The proposed values (p-values) are defined to show the significant difference of 2 performance measures, low p-values indicate that they are statistically different.} The first column of Table \ref{tab:classresults_all} shows our baseline model.

\begin{figure}[t]
\begin{center}
\vspace{-2.5mm}
\includegraphics[width=\linewidth]{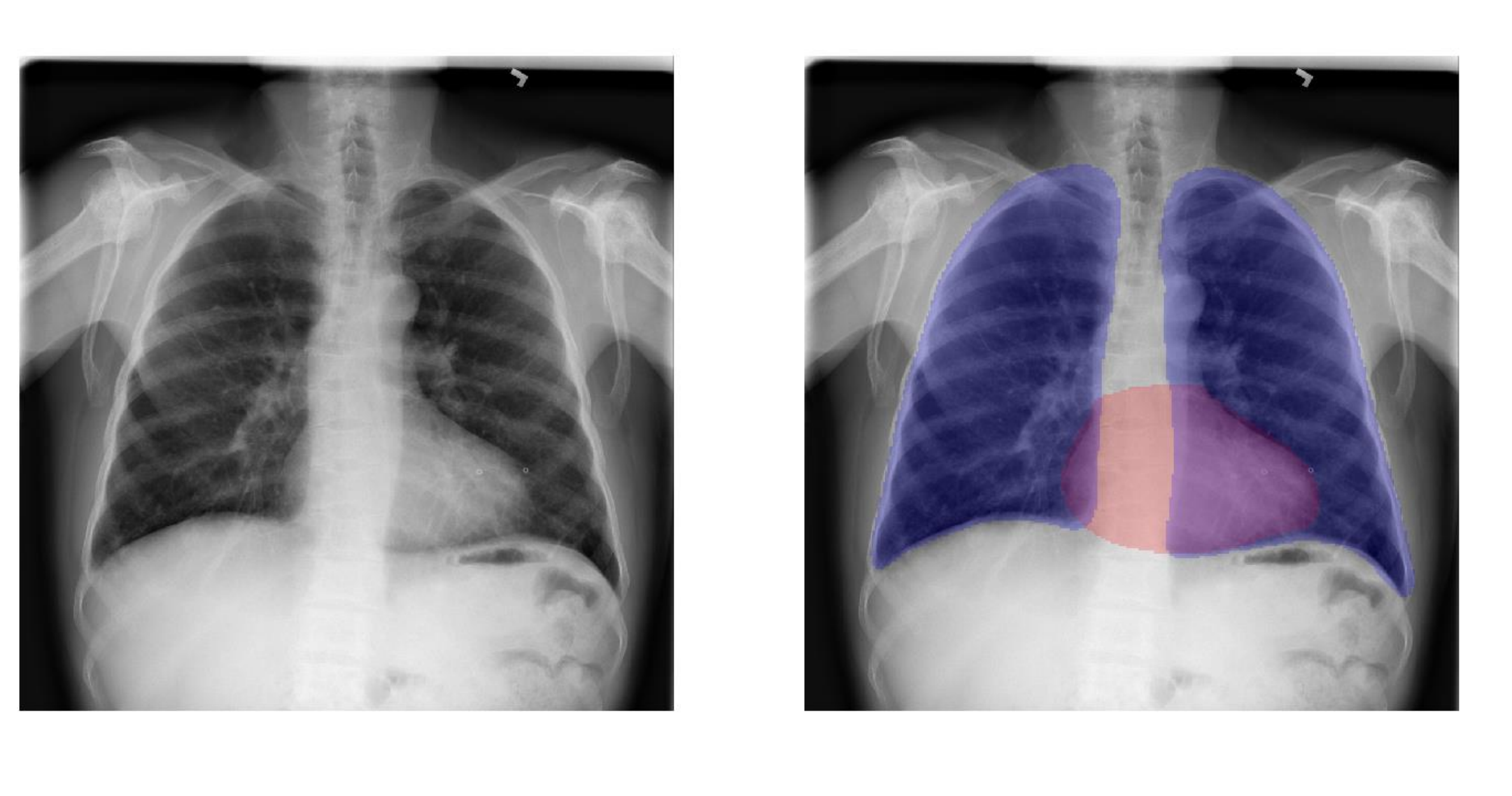}
\caption{Left: Example chest radiograph. Right: Predicted  segmentation masks of lungs (blue) and heart (red).}
\label{fig:segmentation}
\end{center}
\end{figure}

\begin{table}
\small
\parbox{.45\linewidth}{
\caption{Segmentation scores of the multi-task network}
\centering
\begin{tabular}{*3c}
{}   & Dice   & IoU \\
\hline
Heart   &  96.8 & 96.9 \\
Lung   &  98.2 & 94.7\\ 
\hline
\end{tabular}
\label{tab:SegmentationScores2}
}
\hfill
\parbox{.45\linewidth}{
\caption{Spatial classification scores of the multi-task network}
\centering
\begin{tabular}{*2c}
Class &  AUC\\
\hline
Left Lung  & 0.861\\
Right Lung & 0.826\\
Lower & 0.835\\
Lower-middle & 0.734\\
Middle & 0.748\\
Upper-middle & 0.781 \\
Upper & 0.847 \\
Diffuse & 0.846\\
Multiple & 0.709\\ 
\hline
\end{tabular}
\label{tab:SpatialClassification}
}
\end{table}

\begin{table*}
\small
\caption{AUC classification scores for experiments. As test set we use 70\% of the consensus expert reading ChestX-ray14 data and 20\% of the PLCO data. With $+$ we denote the combination between the baseline system and one feature, e.g., ``+norm'' denotes the baseline system with the normalization component.  \ifrevreference\textbf{C2.3: }\fi\changemarker{The columns ``all-noise'', ``all-corr'', and ``all'' denote a combined system with all previous features including noise, correlation regularization, and both, respectively.} The last column shows p-values between the best model (bold values) and the baseline model for each class. \changemarker{The p-values were computed using DeLong's test \citep{PMID:3203132}}}
\vspace{-0.18in}
\begin{center}
\begin{tabular}{ l l c | c | c  c | c  c | c c >{\color{black}}c : c} 

& & & & \multicolumn{2}{c|}{anatomy} & \multicolumn{2}{c|}{regularization} & \multicolumn{3}{c:}{combination}\\

& Abnormalities & \textbf{baseline} & +\textbf{norm} & +\textbf{seg} & +\textbf{loc} & +\textbf{noise} & +\textbf{corr} & \textbf{all-noise} & \textbf{all-corr} & \textbf{all} & p-value\\
\hline
\hline
\parbox[t]{2mm}{\multirow{5}{*}{\rotatebox[origin=c]{90}{\changemarker{Expert Read}}}}
\parbox[t]{2mm}{\multirow{5}{*}{\rotatebox[origin=c]{90}{ChestX-ray14}}} & Effusion & 0.923 & 0.949 & 0.938 & 0.936 & 0.940 & 0.915 & \textbf{0.951} & 0.933 & 0.950 & 0.016\\
& Cardiomegaly & 0.926 & 0.943 & 0.950 & 0.942 & 0.927 & 0.940 & 0.932 & 0.955 & \textbf{0.957} & \changemarker{0.031}\\
& Consolidation & 0.812 & 0.847 & 0.823 & 0.837 & 0.836 & 0.831 & 0.838 & 0.850 & \textbf{0.852} & \changemarker{0.024}\\
& Atelectasis & 0.821 & 0.847 & 0.834 & 0.819 & 0.845 & 0.831 & \textbf{0.851} & 0.840 & 0.848 & 0.045\\
& Mass & 0.804  & 0.815 & 0.805 & 0.842 & 0.829 & 0.815 & 0.830 & \textbf{0.838} & \textbf{0.838} & \changemarker{0.021}\\
\hline

\parbox[t]{2mm}{\multirow{12}{*}{\rotatebox[origin=c]{90}{PLCO}}} & Nodule & 0.809  & 0.817 & 0.815 & 0.821 & 0.817 & 0.816 & 0.821 & 0.825 & \textbf{0.826} & $<0.001$\\
& Mass & 0.837 & 0.851 & 0.858 & 0.844 & 0.846 & 0.845 & \textbf{0.862}& 0.857& 0.859 & 0.011\\
& Granuloma & 0.881 & 0.885 & 0.881 & 0.883 & 0.883 & 0.884 & 0.883 & 0.888 & \textbf{0.889} & $<0.001$\\
& Infiltrate & 0.871 & 0.882 & 0.873 & \textbf{0.885} & 0.878 & 0.876 & 0.878 & 0.879& 0.877 & 0.036\\
& Scaring & 0.842  & 0.845 & 0.846 & 0.845 & 0.847 & 0.847 & 0.847 & 0.850 & \textbf{0.851} & 0.001\\
& Fibrosis & 0.870 & 0.874 & 0.875 & 0.870 & 0.875 & 0.875 & \textbf{0.877} & 0.875 & \textbf{0.877} & 0.015\\
& B./S. Tissue L. & 0.842  & 0.832 & \textbf{0.844} & 0.839 & 0.837 & 0.834 & 0.837 & 0.841& 0.838 & 0.288\\
& Cardiac Abn. & 0.925 & 0.925 & 0.927 & 0.925 & 0.927 & 0.926 & 0.923 & 0.928& \textbf{0.929} & \changemarker{0.201}\\
& COPD & 0.870  & 0.883 & 0.876 & 0.873 & 0.879 & 0.877 & 0.872 & \textbf{0.889}& 0.884& 0.034\\
& Effusion & 0.931 & 0.950 & 0.946 & 0.942 & 0.953 & 0.955 & 0.946 & \textbf{0.956}& 0.955& 0.263\\
& Atelectasis & 0.844 & 0.853 & 0.853 & 0.884 & 0.875 & 0.856 & 0.860 & \textbf{0.889}& 0.888 & 0.043\\
& Hilar Abn. & 0.813 & 0.810 & 0.826 & 0.818 & 0.812 & 0.823 & 0.829 & 0.817& \textbf{0.830} & \changemarker{0.089}\\
\hline

\end{tabular}
\\
\end{center}

\label{tab:classresults_all}
\vspace{-0.2in}
\end{table*}

\begin{figure*}
\begin{center}
\includegraphics[width=5.2in]{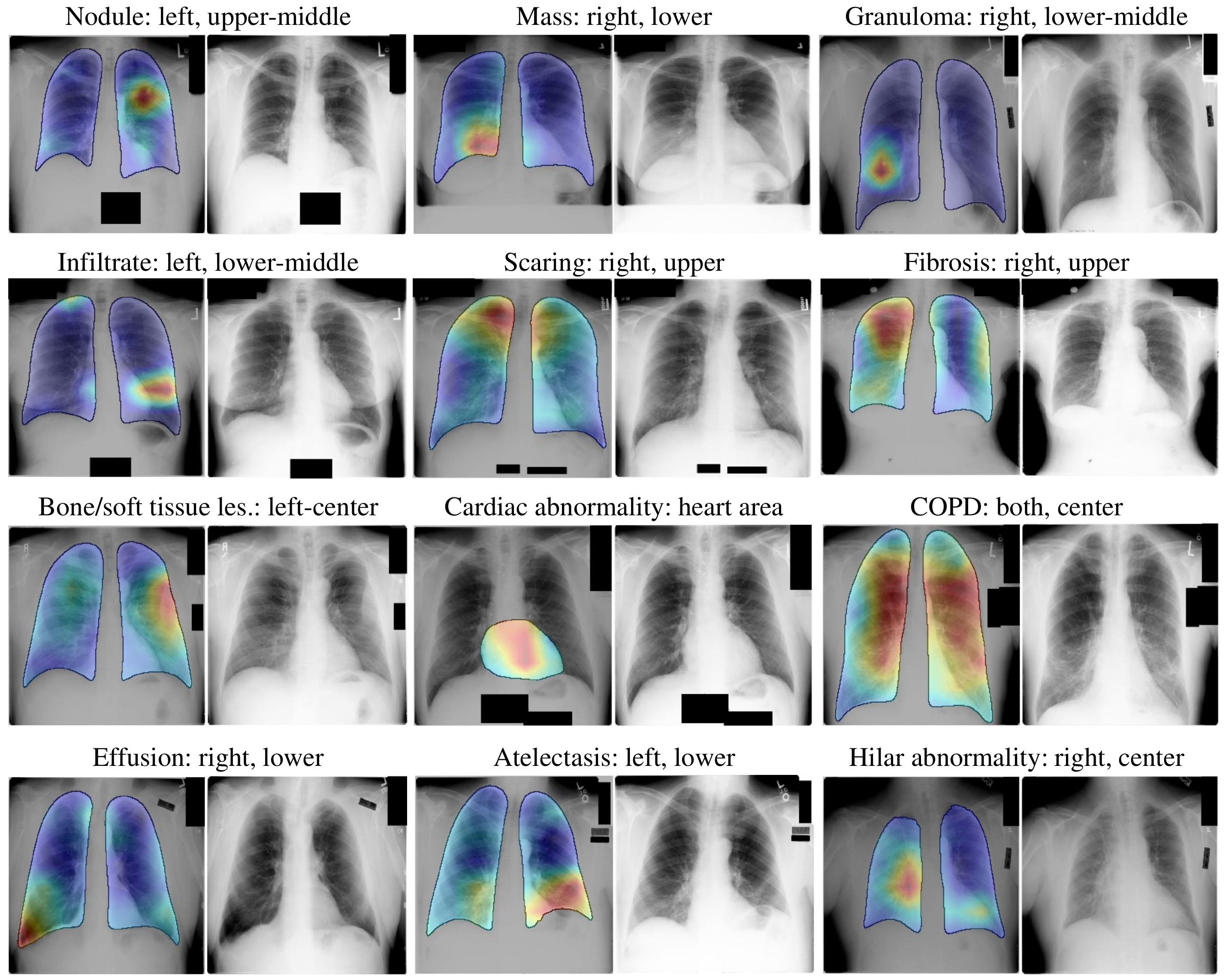}
\vspace{-0.25cm}
\caption{\iffalse\textbf{C2.2:}\fi\textcolor{black}{Heatmap prediction with GradCAM++ \citep{8354201} : Image pairs (left: GradCam++ prediction filtered with lung/heart segmentation masks; right: original image with image normalization) for all 12 PLCO abnormalities from left to right and from top to bottom. All images can be defined as true positive based on a prediction threshold of 0.5.}}
\label{fig:heatmap1}
\end{center}
\end{figure*}

\subsection{Normalization}

 Including the normalization step based on dynamic windowing had a
 two-fold benefit. First, the training time was reduced on average 2-3 times (in terms of average number of epochs). We hypothesize that this is because the normalization ensures images to be more aligned in terms of brightness and contrast, which in some sense simplifies the learning task. Second, this also improved the generalization of the model, and led to a performance increase across all abnormalities, e.g., Effusion (ChestX-ray14) could be improved by 0.026 to 0.949 ($p=0.019$) as can be seen in the Column 2 (\textbf{norm}).  \\

\subsection{Lung and Heart Segmentation}

Column 3 in Table \ref{tab:classresults_all} shows improved classification scores when the network was additionally trained to generate lung and heart segmentation masks (\textbf{seg}). Some abnormalities were significantly improved, e.g., cardiomegaly from 0.929 to 0.950 ($p=0.042$). \par

An example image is visualized in Figure \ref{fig:segmentation} (left). The probabilistic segmentation map is thresholded and overlayed with the image. The red mask defines the heart area, the blue mask indicates the two lungs (right).

\subsection{Spatial Knowledge}
\label{subsec:spatial_knowledge}
We measured the impact of the location labels on the performance of the classification. A performance gain of all abnormalities supported with spatial information could be observed, as can be seen in the fourth column of Table \ref{tab:classresults_all} (\textbf{loc}). Atelectasis (PLCO) improved significantly by 0.040 to an AUC score of 0.884 ($p=0.047$).

In addition, a slight improvement across all abnormalities, can be seen. Thus, we hypothesize that the partial integration of spatial knowledge may also help to improve abnormalities which were not directly supported with spatial information, e.g., due to the abnormality correlation aspect.

\subsection{Regularization}

 We evaluated the noise and correlation experiments on the re-read Chest-Xray14 test set only as the calibration set is derived from this data. In Table \ref{tab:classresults_all}, we can see the performance of both experiments (\textbf{noise}/\textbf{corr}). Adding the regularization terms significantly improved the performance on individual abnormalities. We highlight the results for the consolidation class as the abnormality shows the highest label noise (see Table \ref{tab:sens_spec}) and one of the stronger correlation classes (see Figure \ref{fig:corr}). The performance for this class improved by 0.024 to 0.836 ($p=0.037$) with label noise regularization and the consolidation class by 0.019 to 0.831 ($p=0.044$) with regularization based on label correlation.

\subsection{All-in-One Joint Model}
\textcolor{black}{
For the multi-task learning network, we combined all previous tasks and features in addition to the baseline model:}
\begin{itemize}
    \item[\textcolor{black}{\textbullet}] \textcolor{black}{Normalization of images}
    \item[\textcolor{black}{\textbullet}] \textcolor{black}{Segmentation of lungs and heart}
    \item[\textcolor{black}{\textbullet}] \textcolor{black}{Classification of spatial location of abnormalities}
\end{itemize}

In addition, we used either the label noise regularization (\textbf{all-noise}) or the label correlation regularization (\textbf{all-corr}). The bold values denote the experiment with highest score for each abnormality. The 2 joint experiments share the best scores of most abnormalities. The highest improvement could be achieved on atelectasis. The ChestX-ray14 class improved by 0.030 to 0.851 (\textbf{all-noise}) and the PLCO class by 0.045 to 0.889 (\textbf{all-corr}). If we consider the best values for each abnormality, an average AUC score of 0.880 was reached.

\textbf{Statistical significance: }In order to measure whether the performance scores are statistically significant, we calculated the p-values. In the last column of Table \ref{tab:classresults_all}, p-values can be seen between the reference and the model leading to the highest AUC score (bold values). Except of ``Bone/Soft Tissue Lesion'', ``Cardiac Abnormality'', and ``Effusion'' (PLCO), all abnormalities improved significantly (i.e. $p<0.05$) including the proposed methods. However, we hypothesize that we reach better performance scores evaluating the models on a clean PLCO test set. 

\textbf{Performance of other tasks: }We also evaluated the segmentation of heart and lungs. Table \ref{tab:SegmentationScores2} shows the dice score and Intersection-over-Union (IoU).
\textcolor{black}{The evaluation of the spatial classes can be seen in Table \ref{tab:SpatialClassification}.}

\textbf{Visual Interpretation:} We used GradCAM++ \citep{8354201} to show attention maps. In Figure \ref{fig:heatmap1}, we show image pairs (GradCam++ prediction and original image) for each abnormality of the PLCO dataset. The attention maps are additionally filtered with the help of the lung/heart segmentation masks. For all lung abnormalities, we show attention limited to both lungs and for heart abnormalities, i.e., cardiac abnormality, we show the heart region. The images containing that abnormality were chosen randomly conditioned by a prediction threshold greater than 0.5. All examples show local attention in correspondence to the given abnormality position.

\section{Discussion}

In this study we proposed a multi-task convolutional neural network which outperforms the baseline model in classifying a wide range of abnormalities in chest radiographs. The loss function was designed to deal with noisy labels during training and exploit the correlation between abnormalities. In addition to the classification of the abnormalities, segmentation masks of lungs and heart and approximate spatial classification prediction of the abnormalities shape the architecture of our network. All images were processed with our dynamic normalization strategy as an unknown degree of variation in the chest radiographs exists.\\ 

\begin{table*}[t!]
\small
\begin{center}
\tabcolsep=0.15cm
\begin{tabu}{l c c : c c c c c c c c c c c c c c} 

& Pat. & Re. & Atel. & Card. & Eff. & Inf. & Mass & Nod. & Pneu. & \changemarker{PTX} & Cons. & Ed. & Emph. & Fib. & Pleu. & Hern.\\
\hline
\parbox[t]{2mm}{\multirow{10}{*}{\rotatebox[origin=c]{90}{\changemarker{\textbf{Other Split}}}}} & & &\multicolumn{14}{l}{\citet{DBLP:journals/corr/abs-1803-04565} - DNet}\\
& $\times$ & & 0.826 & 0.911 & 0.885 & 0.716 & 0.854 & 0.774 & 0.765 & 0.872 & 0.806 & 0.892 & 0.925 & 0.820 & 0.785 & 0.941\\
\cline{2-17}
& & & \multicolumn{14}{l}{\citet{fusion_net} - Fusion High-Resolution Network}\\
& $\times$ & & 0.794 & 0.902 & 0.839 & 0.714 & 0.827 & 0.727 & 0.703 & 0.848 & 0.773 & 0.834 & 0.911 & 0.824 & 0.752 & 0.916\\
\cline{2-17}
& & & \multicolumn{14}{l}{\citet{GUAN2018} - Baseline vs. Attention Guided CNN}\\
& & & 0.832 & 0.906 & 0.887 & 0.717 & 0.870 & 0.791 & 0.732 & 0.891 & 0.808 & 0.905 & 0.912 & 0.823 & 0.802 & 0.883\\
& & & 0.853 & 0.939 & 0.903 & 0.754 & 0.902 & 0.828 & 0.774 & 0.921 & 0.842 & 0.924 & 0.932 & 0.864 & 0.837 & 0.921\\
\cline{2-17}
& & &\multicolumn{14}{l}{\citet{10.1371/journal.pmed.1002686} - Radiologist vs. Network}\\
& $\times$ & $\times$ &0.808 & 0.888 & 0.900 & 0.734 & 0.886 & 0.899 & 0.823 & 0.940 & 0.841 & 0.910  & 0.911 & 0.897 & 0.779 & 0.985\\
& $\times$ & $\times$ &0.862 & 0.831 & 0.901 & 0.721 & 0.909 & 0.894 & 0.851 & 0.944 & 0.893 & 0.924 & 0.704 & 0.806 & 0.798 & 0.851\\
\cline{2-17}
& & &\multicolumn{14}{l}{Ours - experiment \textbf{all} based on the re-read ChestX-ray14 subset}\\
& $\times$ & $\times$ & 0.848 & 0.957 & 0.950 & - & 0.838 & - & - & - & 0.852 & - & - & - & - & -\\
\hline

\hline
\hline

\parbox[t]{2mm}{\multirow{7}{*}{\rotatebox[origin=c]{90}{\textbf{\changemarker{Official Split}}}}} & & &\multicolumn{14}{l}{\citet{DBLP:journals/corr/abs-1803-04565} - DNet}\\
& \changemarker{\ifrevreference\,\fi$\times$} & & 0.767 & 0.883 & 0.828 & 0.709 & 0.821 & 0.758 & 0.731 & 0.846 & 0.745 & 0.835 & 0.895 & 0.818 & 0.761 & 0.896\\
\cline{2-17}
& & &\multicolumn{14}{l}{\citet{liu2018sdfn} - Baseline vs. SDFN}\\
& $\times$ & & 0.762 & 0.878 & 0.822 & 0.693 & 0.791 & 0.744 & 0.707 & 0.855 & 0.737 & 0.837 & 0.912 & 0.826 & 0.760 & 0.902\\
& $\times$ & & 0.781 & 0.885 & 0.832 & 0.700 & 0.815 & 0.765 & 0.719 & 0.866 & 0.743 & 0.842 & 0.921 & 0.835 & 0.791 & 0.911\\
\cline{2-17}
\rowfont{\color{black}}& & &\multicolumn{14}{l}{Ours - experiment \textbf{all} based on the official ChestX-ray14 split}\\
\rowfont{\color{black}}& $\times$ & & 0.785 & 0.892 & 0.836 & 0.710 & 0.826 & 0.755 & 0.735 & 0.847 & 0.747 & 0.837 & 0.925 & 0.838 & 0.785 & 0.905\\
\hline

\end{tabu}
\end{center}
\caption{\changemarker{Performance scores (AUC) of different methods based on the ChestX-ray14 dataset. The abbreviated abnormalities are placed in the same order as visualized in Figure \ref{fig:corr} (left). The overall table is separated into methods where the official ChestX-ray14 split and other splits were applied. Furthermore, we highlight key factors of the methods: We separate the methods into random split and patient-wise split (Pat.), additionally, we denote if a subset is re-read by radiologists (Re.). Note that reported performance is influenced by different factors (e.g., data selection, evaluation strategy, reference standard) and therefore, direct comparison between different methods for other splitting strategies is not trivial.} }
\label{tab:comparision}

\end{table*}

We showed that training with prior information helped to achieve a performance gain under label noise. By including label noise ratios and label correlation information the performance increased on the 5 analyzed abnormalities of the ChestX-ray14 dataset. A reading process including 4 board-certified radiologists was performed to calibrate and evaluate our proposed strategies on higher quality reference labels.
The calibration of the loss regularization was based on a small portion, i.e. $<$ 1\% of the whole ChestX-ray14 dataset, and therefore, it is expected that using larger re-read dataset can calibrate the labels more accurately. The regularization predominantly improved abnormalities with stronger noise and correlation, e.g., consolidation from 0.812 AUC to 0.836 (noise) and to 0.831 AUC (correlation). \ifrevreference\textbf{C4.2: }\fi\changemarker{We further show performance scores with noise regularization on the PLCO set. However, the improvement on the ChestX-ray14 dataset exceeded as the PLCO data was not re-read. We hypothesize that a re-reading process of the PLCO may help to increase the performance of the corresponding classes. }\newline
Information about anatomical structures in chest radiographs and location of abnormalities were included. Certain classes significantly improved, e.g., hilar abnormality from 0.813 to 0.826 by including segmentation masks. We hypothesize that the network interpreted the location of the hilar region better as the abnormality is located close to the lung border. The provided nodule location helps to improve the performance from 0.809 to 0.821 as nodules are often hard to find because they appear as a tiny fraction in the images. \newline
Chest radiographs typically appear with high variation in image quality (e.g., contrast ratio, level of exposure, noise level). This is an important challenge, as algorithms should ideally be applicable for images acquired in many institutions. Our dynamic normalization technique was applied to compensate for the strong intensity variation by applying noise filters and linear transformation of image intensities. Evaluated on two different datasets (ChestX-ray14 and PLCO), we showed that our dynamic normalization can be applied to improve the classification performance of the proposed algorithm. \ifrevreference\textbf{C4.2: }\fi\changemarker{We specifically see on the ChestX-ray14 dataset that dynamic normalization contributes mainly with respect to the performance improvement. \citet{ChestXray14Problems2} mentioned that there is a substantially low image quality without any standards, i.e. the image intensities vary significantly across the images. Therefore, the benefit of using the dynamic normalization may be more pronounced on several abnormalities than the benefit of using the proposed regularization. However, there are also other abnormalities where our regularization technique predominated, e.g. mass with 0.815 (normalization) and 0.829 (noise regularization).}  \\

The overall aim of the paper is to build an automated method that can simultaneously predict multiple abnormalities. Based on the complexity of classifying multiple abnormalities with different characteristics, one integrated method cannot help to improve all classes. For that reason, our investigation were focused in searching strategies to improve most of the given abnormalities. In combination of the proposed methods we achieved a significant improvement of the performance for 14 of 17 abnormality classes.

An observer study was performed to evaluate the performance of the NLP labels against radiologists-based labels. While the NLP-based labels maintain high specificity, we observed low sensitivity scores in all relevant abnormalities. This means that using only NLP-based labels may not be ideal for developing deep learning algorithms. The strong label difference between original labels and our radiologist study
disclose the high bias that needs to be taken into account in interpreting the reported performance of algorithms trained using NLP-based labels. Large scale radiologist-based labels remain crucial in this field and, if available, a re-labeling process is necessary in order to uncover label noise. While obtaining radiologists-based labels is crucial, inter and intra-observer variability can still impact the performance. To improve this process, more than 4 radiologists may be considered to further reduce label noise. In this work, the main limitation was the small subset of re-read labels (689 cases), for which 30\% were used to find the parameters for regularization. A higher number of cases for the calibration may further increase the performance. Moreover, we used the original labels for the validation set due to the limited amount of re-read examples. For a more precise validation, the validation set labels also need to be re-read. \newline

\ifrevreference\textbf{C1.9/C4.1: }\fi\changemarker{As we proposed several sub-strategies to increase the performance, our proposed system required a lot of prior work. On the one hand, the data for the auxiliary tasks, i.e., lung/heart segmentation masks and spatial classification labels need to be available to train our multi-task system. On the other hand, the prior re-reading process is expensive, time-consuming, and therefore limited in terms of scalability. However, we show in this work that a small subset is sufficient to obtain a performance gain. Nevertheless, to entirely prevent a prior re-reading process, unsupervised approaches may be considered which adapt the loss depending on noise level probabilities. For example \citet{noise_mixture_model} applied a mixture model to find noisy labels. Some other approaches claim that the mean absolute error function shows more robustness against label noise if weighing training examples with uncertainty differently. \citep{mae_noise, noise_mae_loss}. However, these methods have solely been tested on standard, non-medical datasets. }\\

In the field of chest radiographic abnormality classification, most studies focus on the performance evaluation with the AUC score. The high publication rate helps to analyse proposed methods by comparing the performance scores. However, these performance scores are often directly compared without prior knowledge about strongly influencing factors. In Table \ref{tab:comparision}, we show different studies and highlight essential factors in order to avoid a direct performance comparison between methods. \ifrevreference\textbf{C1.3/C1.8/C2.6: }\fi\changemarker{As highlighted in Section \ref{sec:related}, most groups split data randomly, which leads to unique patients in both training and testing set. For example the ChestX-ray14 dataset has an average of 3.6 images per patient. Therefore, the split should be generated patient-wise which is denoted with ``Pat.'' in Table \ref{tab:comparision}. \newline
\textbf{Official Split:} In this table, we show the performance of the model proposed in \citet{DBLP:journals/corr/abs-1803-04565} using the official patient-wise splits.  In \citet{liu2018sdfn}, both a baseline model and a proposed method, described in Section \ref{sec:related}, were evaluated. The (official) patient-wise split used for both models thus enabling a fair comparison and further enables a comparison with other work. In our approach (last row), we show performance scores of our proposed model \textbf{all}. However, to ensure a fair comparison, we changed to the official ChestX-ray14 split and re-trained the model, in contrast to Table \ref{tab:classresults_all}. We observed improved AUC scores on most abnormalities compared to the other approaches using the official split. \newline
\textbf{Other Split:} In \citet{DBLP:journals/corr/abs-1803-04565}, the same model was used both for the official and another split. Even if both splits are patient-wise, a significant performance difference can be seen across all abnormalities. Thus, a direct comparison of the performance should only be considered if the same split is used since there is a significant performance variability by using different test sets. Another patient-wise split was applied in \citet{fusion_net}. Because of the missing baseline, a comparison to other approaches should be avoided. \citet{GUAN2018} proposed an approach with an additional baseline model on another split, however, the splits were separated randomly.}\newline
Because of the high label noise ratio, \citet{10.1371/journal.pmed.1002686} evaluated their model on a re-read subset (``Re.'' in Table \ref{tab:comparision}). The radiologist performance was compared with the model performance across all 14 abnormalities.
Finally, we show the performance scores of the 5 abnormalities \changemarker{(Table \ref{tab:classresults_all}, \textbf{all})} based on our re-read subset. \newline
\changemarker{In general, we highlight the importance of the applied splitting strategies to enable or avoid performance comparisons.}We see a high performance variance when using different reference labels, e.g, cardiomegaly from 0.831 AUC \citep{10.1371/journal.pmed.1002686} to \changemarker{0.939 AUC \citep{GUAN2018}.}In addition to the increasing availability of chest radiograph datasets for abnormality classification, radiologist studies and different splitting strategies lead to many subsets with different labels. \changemarker{If no re-read subset can be established,} a self-defined baseline model should therefore be a consistent reference to compare the proposed \changemarker{methods in the future.} \\

\section{Conclusion}

We developed a novel method that achieves state-of-the-art performance for the classification of 17 abnormalities based on chest radiographs. A radiologist study enabled an analysis of the accuracy of NLP labels compared to the consensus of 4 board-certified radiologists indicated a large fraction of label noise. We introduced a novel training strategy to handle label noise in the training data. In addition, a regularizer was implemented to exploit the label correlation information during training. The network was expanded to predict segmentation masks of the underlying anatomy as well as the spatial classes for the location of the abnormalities. We implemented a dynamic image normalization technique as the scans are derived from different scanners and post-processing methods. We show that our proposed method can be used to significantly improve the performance of deep learning methods trained on datasets with noisy labels.

\section*{Acknowledgement}
The authors thank the National Cancer Institute (NCI) for access to their data collected by the Prostate, Lung, Colorectal and Ovarian (PLCO) Cancer Screening Trial. The authors thank the National Institutes of Health (NIH) for access to the ChestX-ray14 collection. The statements contained herein are solely those of the authors and do not represent or imply concurrence or endorsement by NCI or NIH.\newline
\textbf{Disclaimer: }The concepts and information presented in this paper are based on research results that are not commercially available. This research did not receive any specific grant from funding agencies in the public, commercial, or not-for-profit sectors.

\bibliographystyle{model2-names.bst}\biboptions{authoryear}
\bibliography{ChestXrayJournal.bbl}

\begin{thebibliography}{67}
\expandafter\ifx\csname natexlab\endcsname\relax\def\natexlab#1{#1}\fi
\providecommand{\url}[1]{\texttt{#1}}
\providecommand{\href}[2]{#2}
\providecommand{\path}[1]{#1}
\providecommand{\DOIprefix}{doi:}
\providecommand{\ArXivprefix}{arXiv:}
\providecommand{\URLprefix}{URL: }
\providecommand{\Pubmedprefix}{pmid:}
\providecommand{\doi}[1]{\href{http://dx.doi.org/#1}{\path{#1}}}
\providecommand{\Pubmed}[1]{\href{pmid:#1}{\path{#1}}}
\providecommand{\bibinfo}[2]{#2}
\ifx\xfnm\relax \def\xfnm[#1]{\unskip,\space#1}\fi
\bibitem[{Amyar et~al.(2020)Amyar, Modzelewski and Ruan}]{multitask}
\bibinfo{author}{Amyar, A.}, \bibinfo{author}{Modzelewski, R.},
  \bibinfo{author}{Ruan, S.}, \bibinfo{year}{2020}.
\newblock \bibinfo{title}{Multi-task deep learning based {CT} imaging analysis
  for {COVID}-19: Classification and segmentation}.
\newblock \DOIprefix\doi{10.1101/2020.04.16.20064709}.
\bibitem[{Arazo et~al.(2019)Arazo, Ortego, Albert, O’Connor and
  McGuinness}]{noise_mixture_model}
\bibinfo{author}{Arazo, E.}, \bibinfo{author}{Ortego, D.},
  \bibinfo{author}{Albert, P.}, \bibinfo{author}{O’Connor, N.},
  \bibinfo{author}{McGuinness, K.}, \bibinfo{year}{2019}.
\newblock \bibinfo{title}{Unsupervised label noise modeling and loss
  correction}.
\bibitem[{Ardila et~al.(2019)Ardila, Kiraly, Bharadwaj, Choi, Reicher, Peng,
  Tse, Etemadi, Ye, Corrado, Naidich and Shetty}]{Ardila2019EndtoendLC}
\bibinfo{author}{Ardila, D.}, \bibinfo{author}{Kiraly, A.P.},
  \bibinfo{author}{Bharadwaj, S.}, \bibinfo{author}{Choi, B.},
  \bibinfo{author}{Reicher, J.J.}, \bibinfo{author}{Peng, L.},
  \bibinfo{author}{Tse, D.}, \bibinfo{author}{Etemadi, M.},
  \bibinfo{author}{Ye, W.}, \bibinfo{author}{Corrado, G.},
  \bibinfo{author}{Naidich, D.P.}, \bibinfo{author}{Shetty, S.},
  \bibinfo{year}{2019}.
\newblock \bibinfo{title}{End-to-end lung cancer screening with
  three-dimensional deep learning on low-dose chest computed tomography}.
\newblock \bibinfo{journal}{Nature Medicine} \bibinfo{volume}{25},
  \bibinfo{pages}{954--961}.
\bibitem[{Ba et~al.(2014)Ba, Mnih and Kavukcuoglu}]{DBLP:journals/corr/BaMK14}
\bibinfo{author}{Ba, J.}, \bibinfo{author}{Mnih, V.},
  \bibinfo{author}{Kavukcuoglu, K.}, \bibinfo{year}{2014}.
\newblock \bibinfo{title}{Multiple object recognition with visual attention}.
\newblock \bibinfo{journal}{CoRR} \bibinfo{volume}{abs/1412.7755}.
\newblock \href{http://arxiv.org/abs/1412.7755}{\tt arXiv:1412.7755}.
\bibitem[{Balabanova et~al.(2005)Balabanova, Coker, Fedorin, Zakharova,
  Plavinskij, Krukov, Atun and Drobniewski}]{agreement_study}
\bibinfo{author}{Balabanova, Y.}, \bibinfo{author}{Coker, R.},
  \bibinfo{author}{Fedorin, I.}, \bibinfo{author}{Zakharova, S.},
  \bibinfo{author}{Plavinskij, S.}, \bibinfo{author}{Krukov, N.},
  \bibinfo{author}{Atun, R.}, \bibinfo{author}{Drobniewski, F.},
  \bibinfo{year}{2005}.
\newblock \bibinfo{title}{Variability in interpretation of chest radiographs
  among russian clinicians and implications for screening programmes:
  Observational study}.
\newblock \bibinfo{journal}{BMJ (Clinical research ed.)} \bibinfo{volume}{331},
  \bibinfo{pages}{379--82}.
\newblock \DOIprefix\doi{10.1136/bmj.331.7513.379}.
\bibitem[{Barbosa~Jr et~al.(2020)Barbosa~Jr, Gefter, Yang, Ghesu, Liu, Mailhe,
  Mansoor, Grbic, Piat, Chabin, Balachandran, Vogt, Ziebandt, Kappler and
  Comaniciu}]{barbosa}
\bibinfo{author}{Barbosa~Jr, E.}, \bibinfo{author}{Gefter, W.},
  \bibinfo{author}{Yang, R.}, \bibinfo{author}{Ghesu, F.},
  \bibinfo{author}{Liu, S.}, \bibinfo{author}{Mailhe, B.},
  \bibinfo{author}{Mansoor, A.}, \bibinfo{author}{Grbic, S.},
  \bibinfo{author}{Piat, S.}, \bibinfo{author}{Chabin, G.},
  \bibinfo{author}{Balachandran, A.}, \bibinfo{author}{Vogt, S.},
  \bibinfo{author}{Ziebandt, V.}, \bibinfo{author}{Kappler, S.},
  \bibinfo{author}{Comaniciu, D.}, \bibinfo{year}{2020}.
\newblock \bibinfo{title}{Automated detection and quantification of {COVID}-19
  airspace disease on chest radiographs: A novel approach achieving
  radiologist-level performance using a cnn trained on digital reconstructed
  radiographs (drrs) from ct-based ground-truth},
  \bibinfo{publisher}{Investigative Radiology}.
\bibitem[{Bayliss et~al.(2005)Bayliss, Ellis and Steiner}]{comorbid}
\bibinfo{author}{Bayliss, E.}, \bibinfo{author}{Ellis, J.},
  \bibinfo{author}{Steiner, J.}, \bibinfo{year}{2005}.
\newblock \bibinfo{title}{Subjective assessments of comorbidity correlate with
  quality of life health outcomes: Initial validation of a comorbidity
  assessment instrument}.
\newblock \bibinfo{journal}{Health and quality of life outcomes}
  \bibinfo{volume}{3}, \bibinfo{pages}{51}.
\newblock \DOIprefix\doi{10.1186/1477-7525-3-51}.
\bibitem[{Brady(2016)}]{article_error}
\bibinfo{author}{Brady, A.P.}, \bibinfo{year}{2016}.
\newblock \bibinfo{title}{Error and discrepancy in radiology: Inevitable or
  avoidable?}
\newblock \bibinfo{journal}{Insights into Imaging} \bibinfo{volume}{8},
  \bibinfo{pages}{171--182}.
\newblock \DOIprefix\doi{10.1007/s13244-016-0534-1}.
\bibitem[{Bruno et~al.(2015)Bruno, Walker and
  Abujudeh}]{doi:10.1148/rg.2015150023}
\bibinfo{author}{Bruno, M.A.}, \bibinfo{author}{Walker, E.A.},
  \bibinfo{author}{Abujudeh, H.H.}, \bibinfo{year}{2015}.
\newblock \bibinfo{title}{Understanding and confronting our mistakes: The
  epidemiology of error in radiology and strategies for error reduction}.
\newblock \bibinfo{journal}{RadioGraphics} \bibinfo{volume}{35},
  \bibinfo{pages}{1668--1676}.
\bibitem[{Cai et~al.(2018)Cai, Lu, Harrison, Shi, Chen and
  Yang}]{attention_mining}
\bibinfo{author}{Cai, J.}, \bibinfo{author}{Lu, L.}, \bibinfo{author}{Harrison,
  A.P.}, \bibinfo{author}{Shi, X.}, \bibinfo{author}{Chen, P.},
  \bibinfo{author}{Yang, L.}, \bibinfo{year}{2018}.
\newblock \bibinfo{title}{Iterative attention mining for weakly supervised
  thoracic disease pattern localization in chest {X-Rays}}, in:
  \bibinfo{editor}{Frangi, A.F.}, \bibinfo{editor}{Schnabel, J.A.},
  \bibinfo{editor}{Davatzikos, C.}, \bibinfo{editor}{Alberola-L{\'o}pez, C.},
  \bibinfo{editor}{Fichtinger, G.} (Eds.), \bibinfo{booktitle}{Medical Image
  Computing and Computer Assisted Intervention -- MICCAI 2018},
  \bibinfo{publisher}{Springer International Publishing},
  \bibinfo{address}{Cham}. pp. \bibinfo{pages}{589--598}.
\bibitem[{{Chattopadhay} et~al.(2018){Chattopadhay}, {Sarkar}, {Howlader} and
  {Balasubramanian}}]{8354201}
\bibinfo{author}{{Chattopadhay}, A.}, \bibinfo{author}{{Sarkar}, A.},
  \bibinfo{author}{{Howlader}, P.}, \bibinfo{author}{{Balasubramanian}, V.N.},
  \bibinfo{year}{2018}.
\newblock \bibinfo{title}{Grad-cam++: Generalized gradient-based visual
  explanations for deep convolutional networks}, in: \bibinfo{booktitle}{2018
  IEEE Winter Conference on Applications of Computer Vision (WACV)}, pp.
  \bibinfo{pages}{839--847}.
\newblock \DOIprefix\doi{10.1109/WACV.2018.00097}.
\bibitem[{DeLong et~al.(1988)DeLong, DeLong and Clarke-Pearson}]{PMID:3203132}
\bibinfo{author}{DeLong, E.}, \bibinfo{author}{DeLong, D.},
  \bibinfo{author}{Clarke-Pearson, D.}, \bibinfo{year}{1988}.
\newblock \bibinfo{title}{Comparing the areas under two or more correlated
  receiver operating characteristic curves: a nonparametric approach}.
\newblock \bibinfo{journal}{Biometrics} \bibinfo{volume}{44},
  \bibinfo{pages}{837--845}.
\newblock \DOIprefix\doi{10.2307/2531595}.
\bibitem[{{Dippel} et~al.(2002){Dippel}, {Stahl}, {Wiemker} and
  {Blaffert}}]{1000258}
\bibinfo{author}{{Dippel}, S.}, \bibinfo{author}{{Stahl}, M.},
  \bibinfo{author}{{Wiemker}, R.}, \bibinfo{author}{{Blaffert}, T.},
  \bibinfo{year}{2002}.
\newblock \bibinfo{title}{Multiscale contrast enhancement for radiographies:
  Laplacian pyramid versus fast wavelet transform}.
\newblock \bibinfo{journal}{IEEE Transactions on Medical Imaging}
  \bibinfo{volume}{21}, \bibinfo{pages}{343--353}.
\bibitem[{Dou et~al.(2020)Dou, Warfield and Gholipour}]{label_noise_overview}
\bibinfo{author}{Dou, H.}, \bibinfo{author}{Warfield, S.},
  \bibinfo{author}{Gholipour, A.}, \bibinfo{year}{2020}.
\newblock \bibinfo{title}{Deep learning with noisy labels: Exploring techniques
  and remedies in medical image analysis}.
\newblock \bibinfo{journal}{Medical Image Analysis} \bibinfo{volume}{65},
  \bibinfo{pages}{101759}.
\newblock \DOIprefix\doi{10.1016/j.media.2020.101759}.
\bibitem[{Fleishon et~al.(2006)Fleishon, Bhargavan and
  Meghea}]{FLEISHON2006453}
\bibinfo{author}{Fleishon, H.B.}, \bibinfo{author}{Bhargavan, M.},
  \bibinfo{author}{Meghea, C.}, \bibinfo{year}{2006}.
\newblock \bibinfo{title}{Radiologists’ reading times using {PACS} and using
  films: One practice’s experience}.
\newblock \bibinfo{journal}{Academic Radiology} \bibinfo{volume}{13},
  \bibinfo{pages}{453 -- 460}.
\bibitem[{Gaál et~al.(2020)Gaál, Maga and Lukács}]{lung_seg_critic}
\bibinfo{author}{Gaál, G.}, \bibinfo{author}{Maga, B.},
  \bibinfo{author}{Lukács, A.}, \bibinfo{year}{2020}.
\newblock \bibinfo{title}{Attention {U}-net based adversarial architectures for
  chest {X-ray} lung segmentation}.
\bibitem[{Gal and Ghahramani(2016)}]{pmlr-v48-gal16}
\bibinfo{author}{Gal, Y.}, \bibinfo{author}{Ghahramani, Z.},
  \bibinfo{year}{2016}.
\newblock \bibinfo{title}{Dropout as a bayesian approximation: Representing
  model uncertainty in deep learning}, in: \bibinfo{editor}{Balcan, M.F.},
  \bibinfo{editor}{Weinberger, K.Q.} (Eds.), \bibinfo{booktitle}{Proceedings of
  The 33rd International Conference on Machine Learning},
  \bibinfo{publisher}{PMLR}, \bibinfo{address}{New York, New York, USA}. pp.
  \bibinfo{pages}{1050--1059}.
\newblock \URLprefix \url{http://proceedings.mlr.press/v48/gal16.html}.
\bibitem[{Ghamrawi and McCallum(2005)}]{Ghamrawi:2005:CMC:1099554.1099591}
\bibinfo{author}{Ghamrawi, N.}, \bibinfo{author}{McCallum, A.},
  \bibinfo{year}{2005}.
\newblock \bibinfo{title}{Collective multi-label classification}, in:
  \bibinfo{booktitle}{Proceedings of the 14th ACM International Conference on
  Information and Knowledge Management}, \bibinfo{publisher}{ACM},
  \bibinfo{address}{New York, NY, USA}. pp. \bibinfo{pages}{195--200}.
\newblock \DOIprefix\doi{10.1145/1099554.1099591}.
\bibitem[{Ghesu et~al.(2019a)Ghesu, Georgescu, Zheng, Grbic, Maier, Hornegger
  and Comaniciu}]{8187667}
\bibinfo{author}{Ghesu, F.}, \bibinfo{author}{Georgescu, B.},
  \bibinfo{author}{Zheng, Y.}, \bibinfo{author}{Grbic, S.},
  \bibinfo{author}{Maier, A.}, \bibinfo{author}{Hornegger, J.},
  \bibinfo{author}{Comaniciu, D.}, \bibinfo{year}{2019}a.
\newblock \bibinfo{title}{Multi-scale deep reinforcement learning for real-time
  3\uppercase{D}-landmark detection in \uppercase{CT} scans}.
\newblock \bibinfo{journal}{IEEE Transactions on Pattern Analysis and Machine
  Intelligence} \bibinfo{volume}{41}, \bibinfo{pages}{176--189}.
\newblock \DOIprefix\doi{10.1109/TPAMI.2017.2782687}.
\bibitem[{Ghesu et~al.(2019b)Ghesu, Georgescu, Gibson, Guendel, Kalra, Singh,
  Digumarthy, Grbic and Comaniciu}]{10.1007/978-3-030-32226-7_75}
\bibinfo{author}{Ghesu, F.C.}, \bibinfo{author}{Georgescu, B.},
  \bibinfo{author}{Gibson, E.}, \bibinfo{author}{Guendel, S.},
  \bibinfo{author}{Kalra, M.K.}, \bibinfo{author}{Singh, R.},
  \bibinfo{author}{Digumarthy, S.R.}, \bibinfo{author}{Grbic, S.},
  \bibinfo{author}{Comaniciu, D.}, \bibinfo{year}{2019}b.
\newblock \bibinfo{title}{Quantifying and leveraging classification uncertainty
  for chest radiograph assessment}, in: \bibinfo{editor}{Shen, D.},
  \bibinfo{editor}{Liu, T.}, \bibinfo{editor}{Peters, T.M.},
  \bibinfo{editor}{Staib, L.H.}, \bibinfo{editor}{Essert, C.},
  \bibinfo{editor}{Zhou, S.}, \bibinfo{editor}{Yap, P.T.},
  \bibinfo{editor}{Khan, A.} (Eds.), \bibinfo{booktitle}{Medical Image
  Computing and Computer Assisted Intervention -- MICCAI 2019},
  \bibinfo{publisher}{Springer International Publishing},
  \bibinfo{address}{Cham}. pp. \bibinfo{pages}{676--684}.
\bibitem[{Ghesu et~al.(2021)Ghesu, Georgescu, Mansoor, Yoo, Gibson, Vishwanath,
  Balachandran, Balter, Cao, Singh, Digumarthy, Kalra, Grbic and
  Comaniciu}]{GHESU2021101855}
\bibinfo{author}{Ghesu, F.C.}, \bibinfo{author}{Georgescu, B.},
  \bibinfo{author}{Mansoor, A.}, \bibinfo{author}{Yoo, Y.},
  \bibinfo{author}{Gibson, E.}, \bibinfo{author}{Vishwanath, R.},
  \bibinfo{author}{Balachandran, A.}, \bibinfo{author}{Balter, J.M.},
  \bibinfo{author}{Cao, Y.}, \bibinfo{author}{Singh, R.},
  \bibinfo{author}{Digumarthy, S.R.}, \bibinfo{author}{Kalra, M.K.},
  \bibinfo{author}{Grbic, S.}, \bibinfo{author}{Comaniciu, D.},
  \bibinfo{year}{2021}.
\newblock \bibinfo{title}{Quantifying and leveraging predictive uncertainty for
  medical image assessment}.
\newblock \bibinfo{journal}{Medical Image Analysis} \bibinfo{volume}{68},
  \bibinfo{pages}{101855}.
\newblock \DOIprefix\doi{https://doi.org/10.1016/j.media.2020.101855}.
\bibitem[{Gómez et~al.(2020)Gómez, Mesejo, Ibáñez, Valsecchi and
  Cordon}]{multi-organ-seg}
\bibinfo{author}{Gómez, O.}, \bibinfo{author}{Mesejo, P.},
  \bibinfo{author}{Ibáñez, O.}, \bibinfo{author}{Valsecchi, A.},
  \bibinfo{author}{Cordon, O.}, \bibinfo{year}{2020}.
\newblock \bibinfo{title}{Deep architectures for high-resolution multi-organ
  chest {X-ray} image segmentation}.
\newblock \bibinfo{journal}{Neural Computing and Applications}
  \bibinfo{volume}{32}.
\newblock \DOIprefix\doi{10.1007/s00521-019-04532-y}.
\bibitem[{Gohagan et~al.(2000)Gohagan, Prorok, Hayes and
  Kramer}]{gohagan2000prostate}
\bibinfo{author}{Gohagan, J.K.}, \bibinfo{author}{Prorok, P.C.},
  \bibinfo{author}{Hayes, R.B.}, \bibinfo{author}{Kramer, B.S.},
  \bibinfo{year}{2000}.
\newblock \bibinfo{title}{The prostate, lung, colorectal and ovarian
  (\uppercase{PLCO}) cancer screening trial of the national cancer institute:
  History, organization, and status}.
\newblock \bibinfo{journal}{Controlled clinical trials} \bibinfo{volume}{21},
  \bibinfo{pages}{251S--272S}.
\bibitem[{Guan and Huang(2018)}]{GUAN2018}
\bibinfo{author}{Guan, Q.}, \bibinfo{author}{Huang, Y.}, \bibinfo{year}{2018}.
\newblock \bibinfo{title}{Multi-label chest {X-ray} image classification via
  category-wise residual attention learning}.
\newblock \bibinfo{journal}{Pattern Recognition Letters}
  \DOIprefix\doi{https://doi.org/10.1016/j.patrec.2018.10.027}.
\bibitem[{Guan et~al.(2018)Guan, Huang, Zhong, Zheng, Zheng and
  Yang}]{2018arXiv180109927G}
\bibinfo{author}{Guan, Q.}, \bibinfo{author}{Huang, Y.},
  \bibinfo{author}{Zhong, Z.}, \bibinfo{author}{Zheng, Z.},
  \bibinfo{author}{Zheng, L.}, \bibinfo{author}{Yang, Y.},
  \bibinfo{year}{2018}.
\newblock \bibinfo{title}{Diagnose like a radiologist: Attention guided
  convolutional neural network for thorax disease classification}.
\newblock \bibinfo{journal}{CoRR} \bibinfo{volume}{abs/1801.09927}.
\bibitem[{Guan et~al.(2020)Guan, Liang, Zhao, Liang, Chen, Li, Liu, Chen, Tang,
  Wang, Ou, Li, Chen, Sang, Wang, Li, Li, Ou, Cheng, Xiong, Ni, Xiang, Hu, Liu,
  Shan, Lei, Peng, Wei, Liu, Hu, Peng, Wang, Liu, Chen, Li, Zheng, Qiu, Luo,
  Ye, Zhu, Cheng, Ye, Li, Zheng, Zhang, Zhong and He}]{Guan2000547}
\bibinfo{author}{Guan, W.j.}, \bibinfo{author}{Liang, W.h.},
  \bibinfo{author}{Zhao, Y.}, \bibinfo{author}{Liang, H.r.},
  \bibinfo{author}{Chen, Z.s.}, \bibinfo{author}{Li, Y.m.},
  \bibinfo{author}{Liu, X.q.}, \bibinfo{author}{Chen, R.c.},
  \bibinfo{author}{Tang, C.l.}, \bibinfo{author}{Wang, T.},
  \bibinfo{author}{Ou, C.q.}, \bibinfo{author}{Li, L.}, \bibinfo{author}{Chen,
  P.y.}, \bibinfo{author}{Sang, L.}, \bibinfo{author}{Wang, W.},
  \bibinfo{author}{Li, J.f.}, \bibinfo{author}{Li, C.c.}, \bibinfo{author}{Ou,
  L.m.}, \bibinfo{author}{Cheng, B.}, \bibinfo{author}{Xiong, S.},
  \bibinfo{author}{Ni, Z.y.}, \bibinfo{author}{Xiang, J.}, \bibinfo{author}{Hu,
  Y.}, \bibinfo{author}{Liu, L.}, \bibinfo{author}{Shan, H.},
  \bibinfo{author}{Lei, C.l.}, \bibinfo{author}{Peng, Y.x.},
  \bibinfo{author}{Wei, L.}, \bibinfo{author}{Liu, Y.}, \bibinfo{author}{Hu,
  Y.h.}, \bibinfo{author}{Peng, P.}, \bibinfo{author}{Wang, J.m.},
  \bibinfo{author}{Liu, J.y.}, \bibinfo{author}{Chen, Z.}, \bibinfo{author}{Li,
  G.}, \bibinfo{author}{Zheng, Z.j.}, \bibinfo{author}{Qiu, S.q.},
  \bibinfo{author}{Luo, J.}, \bibinfo{author}{Ye, C.j.}, \bibinfo{author}{Zhu,
  S.y.}, \bibinfo{author}{Cheng, L.l.}, \bibinfo{author}{Ye, F.},
  \bibinfo{author}{Li, S.y.}, \bibinfo{author}{Zheng, J.p.},
  \bibinfo{author}{Zhang, N.f.}, \bibinfo{author}{Zhong, N.s.},
  \bibinfo{author}{He, J.x.}, \bibinfo{year}{2020}.
\newblock \bibinfo{title}{Comorbidity and its impact on 1590 patients with
  {COVID}-19 in china: A nationwide analysis}.
\newblock \bibinfo{journal}{European Respiratory Journal}
  \DOIprefix\doi{10.1183/13993003.00547-2020}.
\bibitem[{G{\"u}ndel et~al.(2019)G{\"u}ndel, Grbic, Georgescu, Liu, Maier and
  Comaniciu}]{DBLP:journals/corr/abs-1803-04565}
\bibinfo{author}{G{\"u}ndel, S.}, \bibinfo{author}{Grbic, S.},
  \bibinfo{author}{Georgescu, B.}, \bibinfo{author}{Liu, S.},
  \bibinfo{author}{Maier, A.}, \bibinfo{author}{Comaniciu, D.},
  \bibinfo{year}{2019}.
\newblock \bibinfo{title}{Learning to recognize abnormalities in chest
  \uppercase{X}-rays with location-aware dense networks}, in:
  \bibinfo{booktitle}{Progress in Pattern Recognition, Image Analysis, Computer
  Vision, and Applications}, pp. \bibinfo{pages}{757--765}.
\bibitem[{{Hu} et~al.(2018){Hu}, {Shen} and
  {Sun}}]{DBLP:journals/corr/abs-1709-01507}
\bibinfo{author}{{Hu}, J.}, \bibinfo{author}{{Shen}, L.},
  \bibinfo{author}{{Sun}, G.}, \bibinfo{year}{2018}.
\newblock \bibinfo{title}{Squeeze-and-excitation networks}, in:
  \bibinfo{booktitle}{2018 IEEE/CVF Conference on Computer Vision and Pattern
  Recognition}, pp. \bibinfo{pages}{7132--7141}.
\newblock \DOIprefix\doi{10.1109/CVPR.2018.00745}.
\bibitem[{{Huang} et~al.(2017){Huang}, {Liu}, v.~d. {Maaten} and
  {Weinberger}}]{huang2018archive}
\bibinfo{author}{{Huang}, G.}, \bibinfo{author}{{Liu}, Z.},
  \bibinfo{author}{v.~d. {Maaten}, L.}, \bibinfo{author}{{Weinberger}, K.Q.},
  \bibinfo{year}{2017}.
\newblock \bibinfo{title}{Densely connected convolutional networks}, in:
  \bibinfo{booktitle}{2017 IEEE Conference on Computer Vision and Pattern
  Recognition (CVPR)}, pp. \bibinfo{pages}{2261--2269}.
\newblock \DOIprefix\doi{10.1109/CVPR.2017.243}.
\bibitem[{Huang et~al.(2020)Huang, Lin, Xu, Wang, Bai, Pang and
  Meen}]{fusion_net}
\bibinfo{author}{Huang, Z.}, \bibinfo{author}{Lin, J.}, \bibinfo{author}{Xu,
  L.}, \bibinfo{author}{Wang, H.}, \bibinfo{author}{Bai, T.},
  \bibinfo{author}{Pang, Y.}, \bibinfo{author}{Meen, T.H.},
  \bibinfo{year}{2020}.
\newblock \bibinfo{title}{Fusion high-resolution network for diagnosing
  {ChestX-ray} images}.
\newblock \bibinfo{journal}{Electronics} \bibinfo{volume}{9},
  \bibinfo{pages}{190}.
\newblock \DOIprefix\doi{10.3390/electronics9010190}.
\bibitem[{Irvin et~al.(2019)Irvin, Rajpurkar, Ko, Yu, Ciurea-Ilcus, Chute,
  Marklund, Haghgoo, Ball, Shpanskaya et~al.}]{irvin2019chexpert}
\bibinfo{author}{Irvin, J.}, \bibinfo{author}{Rajpurkar, P.},
  \bibinfo{author}{Ko, M.}, \bibinfo{author}{Yu, Y.},
  \bibinfo{author}{Ciurea-Ilcus, S.}, \bibinfo{author}{Chute, C.},
  \bibinfo{author}{Marklund, H.}, \bibinfo{author}{Haghgoo, B.},
  \bibinfo{author}{Ball, R.}, \bibinfo{author}{Shpanskaya, K.}, et~al.,
  \bibinfo{year}{2019}.
\newblock \bibinfo{title}{{CheXpert:} a large chest radiograph dataset with
  uncertainty labels and expert comparison}, in:
  \bibinfo{booktitle}{Thirty-Third AAAI Conference on Artificial Intelligence}.
\bibitem[{Islam et~al.(2017)Islam, Aowal, Minhaz and
  Ashraf}]{DBLP:journals/corr/IslamAMA17}
\bibinfo{author}{Islam, M.T.}, \bibinfo{author}{Aowal, M.A.},
  \bibinfo{author}{Minhaz, A.T.}, \bibinfo{author}{Ashraf, K.},
  \bibinfo{year}{2017}.
\newblock \bibinfo{title}{Abnormality detection and localization in chest
  \uppercase{X}-rays using deep convolutional neural networks}.
\newblock \bibinfo{journal}{CoRR} \bibinfo{volume}{abs/1705.09850}.
\newblock \href{http://arxiv.org/abs/1705.09850}{\tt arXiv:1705.09850}.
\bibitem[{J\o{}sang(2016)}]{10.5555/3031657}
\bibinfo{author}{J\o{}sang, A.}, \bibinfo{year}{2016}.
\newblock \bibinfo{title}{Subjective Logic: A Formalism for Reasoning Under
  Uncertainty}.
\newblock \bibinfo{edition}{1st} ed., \bibinfo{publisher}{Springer Publishing
  Company, Incorporated}.
\bibitem[{Jusoh(2018)}]{NLPamb}
\bibinfo{author}{Jusoh, S.}, \bibinfo{year}{2018}.
\newblock \bibinfo{title}{A study on nlp applications and ambiguity problems}.
\newblock \bibinfo{journal}{Journal of Theoretical and Applied Information
  Technology} \bibinfo{volume}{96}, \bibinfo{pages}{1486--1499}.
\bibitem[{Kholiavchenko et~al.(2020)Kholiavchenko, Sirazitdinov, Kubrak,
  Badrutdinova, Kuleev, Yuan, Vrtovec and Ibragimov}]{contour_seg}
\bibinfo{author}{Kholiavchenko, M.}, \bibinfo{author}{Sirazitdinov, I.},
  \bibinfo{author}{Kubrak, K.}, \bibinfo{author}{Badrutdinova, R.},
  \bibinfo{author}{Kuleev, R.}, \bibinfo{author}{Yuan, Y.},
  \bibinfo{author}{Vrtovec, T.}, \bibinfo{author}{Ibragimov, B.},
  \bibinfo{year}{2020}.
\newblock \bibinfo{title}{Contour-aware multi-label chest {X-ray} organ
  segmentation}.
\newblock \bibinfo{journal}{International Journal of Computer Assisted
  Radiology and Surgery} \bibinfo{volume}{15}.
\newblock \DOIprefix\doi{10.1007/s11548-019-02115-9}.
\bibitem[{Kingma and Ba(2015)}]{adam}
\bibinfo{author}{Kingma, D.P.}, \bibinfo{author}{Ba, J.}, \bibinfo{year}{2015}.
\newblock \bibinfo{title}{Adam: {A} method for stochastic optimization}, in:
  \bibinfo{editor}{Bengio, Y.}, \bibinfo{editor}{LeCun, Y.} (Eds.),
  \bibinfo{booktitle}{3rd International Conference on Learning Representations,
  {ICLR} 2015, San Diego, CA, USA, May 7-9, 2015, Conference Track
  Proceedings}.
\bibitem[{Lakshminarayanan et~al.(2017)Lakshminarayanan, Pritzel and
  Blundell}]{10.5555/3295222.3295387}
\bibinfo{author}{Lakshminarayanan, B.}, \bibinfo{author}{Pritzel, A.},
  \bibinfo{author}{Blundell, C.}, \bibinfo{year}{2017}.
\newblock \bibinfo{title}{Simple and scalable predictive uncertainty estimation
  using deep ensembles}, in: \bibinfo{booktitle}{Proceedings of the 31st
  International Conference on Neural Information Processing Systems},
  \bibinfo{publisher}{Curran Associates Inc.}, \bibinfo{address}{Red Hook, NY,
  USA}. p. \bibinfo{pages}{6405–6416}.
\bibitem[{{Larrazabal} et~al.(2020){Larrazabal}, {Martínez}, {Glocker} and
  {Ferrante}}]{9126830}
\bibinfo{author}{{Larrazabal}, A.J.}, \bibinfo{author}{{Martínez}, C.},
  \bibinfo{author}{{Glocker}, B.}, \bibinfo{author}{{Ferrante}, E.},
  \bibinfo{year}{2020}.
\newblock \bibinfo{title}{{Post-DAE}: Anatomically plausible segmentation via
  post-processing with denoising autoencoders}.
\newblock \bibinfo{journal}{IEEE Transactions on Medical Imaging}
  \bibinfo{volume}{39}, \bibinfo{pages}{3813--3820}.
\newblock \DOIprefix\doi{10.1109/TMI.2020.3005297}.
\bibitem[{Li et~al.(2016)Li, Cao, Zhao and Wang}]{Li2016PulmonaryNC}
\bibinfo{author}{Li, W.}, \bibinfo{author}{Cao, P.}, \bibinfo{author}{Zhao,
  D.}, \bibinfo{author}{Wang, J.}, \bibinfo{year}{2016}.
\newblock \bibinfo{title}{Pulmonary nodule classification with deep
  convolutional neural networks on computed tomography images}.
\newblock \bibinfo{journal}{Computational and Mathematical Methods in Medicine}
  \bibinfo{volume}{2016}, \bibinfo{pages}{1--7}.
\bibitem[{Li et~al.(2018)Li, Wang, Han, Xue, Wei, Li and
  Fei-Fei}]{DBLP:journals/corr/abs-1711-06373}
\bibinfo{author}{Li, Z.}, \bibinfo{author}{Wang, C.}, \bibinfo{author}{Han,
  M.}, \bibinfo{author}{Xue, Y.}, \bibinfo{author}{Wei, W.},
  \bibinfo{author}{Li, L.}, \bibinfo{author}{Fei-Fei, L.},
  \bibinfo{year}{2018}.
\newblock \bibinfo{title}{Thoracic disease identification and localization with
  limited supervision}, in: \bibinfo{booktitle}{2018 IEEE/CVF Conference on
  Computer Vision and Pattern Recognition (CVPR)}, pp.
  \bibinfo{pages}{8290--8299}.
\newblock \DOIprefix\doi{10.1109/CVPR.2018.00865}.
\bibitem[{Liu et~al.(2019)Liu, Wang, Nan, Jin, Wang and Pu}]{liu2018sdfn}
\bibinfo{author}{Liu, H.}, \bibinfo{author}{Wang, L.}, \bibinfo{author}{Nan,
  Y.}, \bibinfo{author}{Jin, F.}, \bibinfo{author}{Wang, Q.},
  \bibinfo{author}{Pu, J.}, \bibinfo{year}{2019}.
\newblock \bibinfo{title}{{SDFN}: Segmentation-based deep fusion network for
  thoracic disease classification in chest {X-ray} images}.
\newblock \bibinfo{journal}{Computerized Medical Imaging and Graphics}
  \bibinfo{volume}{75}, \bibinfo{pages}{66--73}.
\newblock \DOIprefix\doi{https://doi.org/10.1016/j.compmedimag.2019.05.005}.
\bibitem[{Molchanov et~al.(2017)Molchanov, Ashukha and
  Vetrov}]{10.5555/3305890.3305939}
\bibinfo{author}{Molchanov, D.}, \bibinfo{author}{Ashukha, A.},
  \bibinfo{author}{Vetrov, D.}, \bibinfo{year}{2017}.
\newblock \bibinfo{title}{Variational dropout sparsifies deep neural networks},
  in: \bibinfo{booktitle}{Proceedings of the 34th International Conference on
  Machine Learning - Volume 70}, \bibinfo{publisher}{JMLR.org}. p.
  \bibinfo{pages}{2498–2507}.
\bibitem[{Oakden-Rayner(2017)}]{ChestXray14Problems}
\bibinfo{author}{Oakden-Rayner, L.}, \bibinfo{year}{2017}.
\newblock \bibinfo{title}{Exploring the chestxray14 dataset: problems}.
\newblock
  \bibinfo{howpublished}{\url{https://lukeoakdenrayner.wordpress.com/2017/12/18/the-chestxray14-dataset-problems/}}.
\newblock \bibinfo{note}{Accessed: 2017-12-18}.
\bibitem[{Oakden-Rayner(2019)}]{ChestXray14Problems2}
\bibinfo{author}{Oakden-Rayner, L.}, \bibinfo{year}{2019}.
\newblock \bibinfo{title}{Half a million x-rays! first impressions of the
  {Stanford} and {MIT} chest x-ray datasets}.
\newblock
  \bibinfo{howpublished}{\url{https://lukeoakdenrayner.wordpress.com/2019/02/25/half-a-million-x-rays-first-impressions-of-the-stanford
  \\-and-mit-chest-x-ray-datasets/}}.
\newblock \bibinfo{note}{Accessed: 2019-02-25}.
\bibitem[{Philipsen et~al.(2015)Philipsen, Maduskar, Hogeweg, Melendez,
  S{\'a}nchez and van Ginneken}]{Philipsen2015LocalizedEN}
\bibinfo{author}{Philipsen, R.H.H.M.}, \bibinfo{author}{Maduskar, P.},
  \bibinfo{author}{Hogeweg, L.}, \bibinfo{author}{Melendez, J.},
  \bibinfo{author}{S{\'a}nchez, C.I.}, \bibinfo{author}{van Ginneken, B.},
  \bibinfo{year}{2015}.
\newblock \bibinfo{title}{Localized energy-based normalization of medical
  images: Application to chest radiography}.
\newblock \bibinfo{journal}{IEEE Transactions on Medical Imaging}
  \bibinfo{volume}{34}, \bibinfo{pages}{1965--1975}.
\bibitem[{Rajkomar et~al.(2018)Rajkomar, Oren, Chen, Dai, Hajaj, Hardt, Liu,
  Liu, Marcus, Sun, Sundberg, Yee, Zhang, Zhang, Flores, Duggan, Irvine, Le,
  Litsch, Mossin, Tansuwan, Wang, Wexler, Wilson, Ludwig, Volchenboum, Chou,
  Pearson, Madabushi, Shah, Butte, Howell, Cui, Corrado and
  Dean}]{Rajkomar_2018}
\bibinfo{author}{Rajkomar, A.}, \bibinfo{author}{Oren, E.},
  \bibinfo{author}{Chen, K.}, \bibinfo{author}{Dai, A.M.},
  \bibinfo{author}{Hajaj, N.}, \bibinfo{author}{Hardt, M.},
  \bibinfo{author}{Liu, P.J.}, \bibinfo{author}{Liu, X.},
  \bibinfo{author}{Marcus, J.}, \bibinfo{author}{Sun, M.},
  \bibinfo{author}{Sundberg, P.}, \bibinfo{author}{Yee, H.},
  \bibinfo{author}{Zhang, K.}, \bibinfo{author}{Zhang, Y.},
  \bibinfo{author}{Flores, G.}, \bibinfo{author}{Duggan, G.E.},
  \bibinfo{author}{Irvine, J.}, \bibinfo{author}{Le, Q.},
  \bibinfo{author}{Litsch, K.}, \bibinfo{author}{Mossin, A.},
  \bibinfo{author}{Tansuwan, J.}, \bibinfo{author}{Wang, D.},
  \bibinfo{author}{Wexler, J.}, \bibinfo{author}{Wilson, J.},
  \bibinfo{author}{Ludwig, D.}, \bibinfo{author}{Volchenboum, S.L.},
  \bibinfo{author}{Chou, K.}, \bibinfo{author}{Pearson, M.},
  \bibinfo{author}{Madabushi, S.}, \bibinfo{author}{Shah, N.H.},
  \bibinfo{author}{Butte, A.J.}, \bibinfo{author}{Howell, M.D.},
  \bibinfo{author}{Cui, C.}, \bibinfo{author}{Corrado, G.S.},
  \bibinfo{author}{Dean, J.}, \bibinfo{year}{2018}.
\newblock \bibinfo{title}{Scalable and accurate deep learning with electronic
  health records}.
\newblock \bibinfo{journal}{npj Digital Medicine} \bibinfo{volume}{1}.
\bibitem[{Rajpurkar et~al.(2018)Rajpurkar, Irvin, Ball, Zhu, Yang, Mehta, Duan,
  Ding, Bagul, Langlotz, Patel, Yeom, Shpanskaya, Blankenberg, Seekins,
  Amrhein, Mong, Halabi, Zucker, Ng and Lungren}]{10.1371/journal.pmed.1002686}
\bibinfo{author}{Rajpurkar, P.}, \bibinfo{author}{Irvin, J.},
  \bibinfo{author}{Ball, R.L.}, \bibinfo{author}{Zhu, K.},
  \bibinfo{author}{Yang, B.}, \bibinfo{author}{Mehta, H.},
  \bibinfo{author}{Duan, T.}, \bibinfo{author}{Ding, D.},
  \bibinfo{author}{Bagul, A.}, \bibinfo{author}{Langlotz, C.P.},
  \bibinfo{author}{Patel, B.N.}, \bibinfo{author}{Yeom, K.W.},
  \bibinfo{author}{Shpanskaya, K.}, \bibinfo{author}{Blankenberg, F.G.},
  \bibinfo{author}{Seekins, J.}, \bibinfo{author}{Amrhein, T.J.},
  \bibinfo{author}{Mong, D.A.}, \bibinfo{author}{Halabi, S.S.},
  \bibinfo{author}{Zucker, E.J.}, \bibinfo{author}{Ng, A.Y.},
  \bibinfo{author}{Lungren, M.P.}, \bibinfo{year}{2018}.
\newblock \bibinfo{title}{Deep learning for chest radiograph diagnosis: A
  retrospective comparison of the \uppercase{C}he\uppercase{XN}e\uppercase{X}t
  algorithm to practicing radiologists}.
\newblock \bibinfo{journal}{PLOS Medicine} \bibinfo{volume}{15},
  \bibinfo{pages}{1--17}.
\newblock \DOIprefix\doi{10.1371/journal.pmed.1002686}.
\bibitem[{Rajpurkar et~al.(2017)Rajpurkar, Irvin, Zhu, Yang, Mehta, Duan, Ding,
  Bagul, Langlotz, Shpanskaya et~al.}]{rajpurkar2017chexnet}
\bibinfo{author}{Rajpurkar, P.}, \bibinfo{author}{Irvin, J.},
  \bibinfo{author}{Zhu, K.}, \bibinfo{author}{Yang, B.},
  \bibinfo{author}{Mehta, H.}, \bibinfo{author}{Duan, T.},
  \bibinfo{author}{Ding, D.}, \bibinfo{author}{Bagul, A.},
  \bibinfo{author}{Langlotz, C.}, \bibinfo{author}{Shpanskaya, K.}, et~al.,
  \bibinfo{year}{2017}.
\newblock \bibinfo{title}{{CheXNet}: Radiologist-level pneumonia detection on
  chest \uppercase{X}-rays with deep learning}
  \bibinfo{volume}{abs/1711.05225}.
\bibitem[{Rolnick et~al.(2017)Rolnick, Veit, Belongie and
  Shavit}]{label_noise_robust}
\bibinfo{author}{Rolnick, D.}, \bibinfo{author}{Veit, A.},
  \bibinfo{author}{Belongie, S.}, \bibinfo{author}{Shavit, N.},
  \bibinfo{year}{2017}.
\newblock \bibinfo{title}{Deep learning is robust to massive label noise} .
\bibitem[{Rubin et~al.(2018)Rubin, Sanghavi, Zhao, Lee, Qadir and
  Xu{-}Wilson}]{DBLP:journals/corr/abs-1804-07839}
\bibinfo{author}{Rubin, J.}, \bibinfo{author}{Sanghavi, D.},
  \bibinfo{author}{Zhao, C.}, \bibinfo{author}{Lee, K.},
  \bibinfo{author}{Qadir, A.}, \bibinfo{author}{Xu{-}Wilson, M.},
  \bibinfo{year}{2018}.
\newblock \bibinfo{title}{Large scale automated reading of frontal and lateral
  chest {X-Rays} using dual convolutional neural networks}.
\newblock \bibinfo{journal}{CoRR} \bibinfo{volume}{abs/1804.07839}.
\bibitem[{Rusiecki(2019)}]{cce_noise}
\bibinfo{author}{Rusiecki, A.}, \bibinfo{year}{2019}.
\newblock \bibinfo{title}{Trimmed Robust Loss Function for Training Deep Neural
  Networks with Label Noise}.
\newblock pp. \bibinfo{pages}{215--222}.
\newblock \DOIprefix\doi{10.1007/978-3-030-20912-4_21}.
\bibitem[{Russakovsky et~al.(2015)Russakovsky, Deng, Su, Krause, Satheesh, Ma,
  Huang, Karpathy, Khosla, Bernstein, Berg and Fei-Fei}]{ImageNet}
\bibinfo{author}{Russakovsky, O.}, \bibinfo{author}{Deng, J.},
  \bibinfo{author}{Su, H.}, \bibinfo{author}{Krause, J.},
  \bibinfo{author}{Satheesh, S.}, \bibinfo{author}{Ma, S.},
  \bibinfo{author}{Huang, Z.}, \bibinfo{author}{Karpathy, A.},
  \bibinfo{author}{Khosla, A.}, \bibinfo{author}{Bernstein, M.},
  \bibinfo{author}{Berg, A.C.}, \bibinfo{author}{Fei-Fei, L.},
  \bibinfo{year}{2015}.
\newblock \bibinfo{title}{{ImageNet} large scale visual recognition challenge}.
\newblock \bibinfo{journal}{IJCV} \bibinfo{volume}{115},
  \bibinfo{pages}{211--252}.
\bibitem[{Selvan et~al.(2020)Selvan, Dam, Detlefsen, Rischel, Sheng, Nielsen
  and Pai}]{dense_segment}
\bibinfo{author}{Selvan, R.}, \bibinfo{author}{Dam, E.},
  \bibinfo{author}{Detlefsen, N.}, \bibinfo{author}{Rischel, S.},
  \bibinfo{author}{Sheng, K.}, \bibinfo{author}{Nielsen, M.},
  \bibinfo{author}{Pai, A.}, \bibinfo{year}{2020}.
\newblock \bibinfo{title}{Lung segmentation from chest {X-rays} using
  variational data imputation}.
\bibitem[{Shen and Gao(2018)}]{10.1007/978-3-030-00919-9_45}
\bibinfo{author}{Shen, Y.}, \bibinfo{author}{Gao, M.}, \bibinfo{year}{2018}.
\newblock \bibinfo{title}{Dynamic routing on deep neural network for thoracic
  disease classification and sensitive area localization}, in:
  \bibinfo{editor}{Shi, Y.}, \bibinfo{editor}{Suk, H.I.}, \bibinfo{editor}{Liu,
  M.} (Eds.), \bibinfo{booktitle}{Machine Learning in Medical Imaging},
  \bibinfo{publisher}{Springer International Publishing},
  \bibinfo{address}{Cham}. pp. \bibinfo{pages}{389--397}.
\bibitem[{Tang et~al.(2018)Tang, Wang, Harrison, Lu, Xiao and
  Summers}]{10.1007/978-3-030-00919-9_29}
\bibinfo{author}{Tang, Y.}, \bibinfo{author}{Wang, X.},
  \bibinfo{author}{Harrison, A.P.}, \bibinfo{author}{Lu, L.},
  \bibinfo{author}{Xiao, J.}, \bibinfo{author}{Summers, R.M.},
  \bibinfo{year}{2018}.
\newblock \bibinfo{title}{Attention-guided curriculum learning for weakly
  supervised classification and localization of thoracic diseases on chest
  radiographs}, in: \bibinfo{editor}{Shi, Y.}, \bibinfo{editor}{Suk, H.I.},
  \bibinfo{editor}{Liu, M.} (Eds.), \bibinfo{booktitle}{Machine Learning in
  Medical Imaging}, \bibinfo{publisher}{Springer International Publishing},
  \bibinfo{address}{Cham}. pp. \bibinfo{pages}{249--258}.
\bibitem[{Van~Eeden et~al.(2012)Van~Eeden, Leipsic, Paul~Man and
  Sin}]{doi:10.1164/rccm.201203-0455PP}
\bibinfo{author}{Van~Eeden, S.}, \bibinfo{author}{Leipsic, J.},
  \bibinfo{author}{Paul~Man, S.F.}, \bibinfo{author}{Sin, D.D.},
  \bibinfo{year}{2012}.
\newblock \bibinfo{title}{The relationship between lung inflammation and
  cardiovascular disease}.
\newblock \bibinfo{journal}{American Journal of Respiratory and Critical Care
  Medicine} \bibinfo{volume}{186}, \bibinfo{pages}{11--16}.
\newblock \DOIprefix\doi{10.1164/rccm.201203-0455PP},
  \href{http://arxiv.org/abs/https://doi.org/10.1164/rccm.201203-0455PP}{\tt
  arXiv:https://doi.org/10.1164/rccm.201203-0455PP}. \bibinfo{note}{pMID:
  22538803}.
\bibitem[{{Wang} et~al.(2020){Wang}, {Jia}, {Lu} and {Xia}}]{ChestNet}
\bibinfo{author}{{Wang}, H.}, \bibinfo{author}{{Jia}, H.},
  \bibinfo{author}{{Lu}, L.}, \bibinfo{author}{{Xia}, Y.},
  \bibinfo{year}{2020}.
\newblock \bibinfo{title}{{Thorax-Net}: An attention regularized deep neural
  network for classification of thoracic diseases on chest radiography}.
\newblock \bibinfo{journal}{IEEE Journal of Biomedical and Health Informatics}
  \bibinfo{volume}{24}, \bibinfo{pages}{475--485}.
\newblock \DOIprefix\doi{10.1109/JBHI.2019.2928369}.
\bibitem[{Wang et~al.(2021)Wang, Wang, Qin, Zhang, Li and
  Xia}]{TripleAttention}
\bibinfo{author}{Wang, H.}, \bibinfo{author}{Wang, S.}, \bibinfo{author}{Qin,
  Z.}, \bibinfo{author}{Zhang, Y.}, \bibinfo{author}{Li, R.},
  \bibinfo{author}{Xia, Y.}, \bibinfo{year}{2021}.
\newblock \bibinfo{title}{Triple attention learning for classification of 14
  thoracic diseases using chest radiography}.
\newblock \bibinfo{journal}{Medical Image Analysis} \bibinfo{volume}{67},
  \bibinfo{pages}{101846}.
\newblock \DOIprefix\doi{10.1016/j.media.2020.101846}.
\bibitem[{Wang et~al.(2019)Wang, Kodirov, Hua and Robertson}]{mae_noise}
\bibinfo{author}{Wang, X.}, \bibinfo{author}{Kodirov, E.},
  \bibinfo{author}{Hua, Y.}, \bibinfo{author}{Robertson, N.},
  \bibinfo{year}{2019}.
\newblock \bibinfo{title}{{IMAE} for noise-robust learning: Mean absolute error
  does not treat examples equally and gradient magnitude’s variance matters}.
\bibitem[{{Wang} et~al.(2017){Wang}, {Peng}, {Lu}, {Lu}, {Bagheri} and
  {Summers}}]{wang2017chestx}
\bibinfo{author}{{Wang}, X.}, \bibinfo{author}{{Peng}, Y.},
  \bibinfo{author}{{Lu}, L.}, \bibinfo{author}{{Lu}, Z.},
  \bibinfo{author}{{Bagheri}, M.}, \bibinfo{author}{{Summers}, R.M.},
  \bibinfo{year}{2017}.
\newblock \bibinfo{title}{Chest\uppercase{x}-ray8: Hospital-scale chest {X-ray}
  database and benchmarks on weakly-supervised classification and localization
  of common thorax diseases}, in: \bibinfo{booktitle}{2017 IEEE Conference on
  Computer Vision and Pattern Recognition (CVPR)}, pp.
  \bibinfo{pages}{3462--3471}.
\bibitem[{Yan et~al.(2018)Yan, Yao, Li, Xu and
  Huang}]{DBLP:journals/corr/abs-1807-06067}
\bibinfo{author}{Yan, C.}, \bibinfo{author}{Yao, J.}, \bibinfo{author}{Li, R.},
  \bibinfo{author}{Xu, Z.}, \bibinfo{author}{Huang, J.}, \bibinfo{year}{2018}.
\newblock \bibinfo{title}{Weakly supervised deep learning for thoracic disease
  classification and localization on chest \uppercase{X}-rays}, in:
  \bibinfo{booktitle}{Proceedings of the 2018 ACM International Conference on
  Bioinformatics, Computational Biology, and Health Informatics}, pp.
  \bibinfo{pages}{103--110}.
\newblock \DOIprefix\doi{10.1145/3233547.3233573}.
\bibitem[{Yao et~al.(2017)Yao, Poblenz, Dagunts, Covington, Bernard and
  Lyman}]{yao2017learning}
\bibinfo{author}{Yao, L.}, \bibinfo{author}{Poblenz, E.},
  \bibinfo{author}{Dagunts, D.}, \bibinfo{author}{Covington, B.},
  \bibinfo{author}{Bernard, D.}, \bibinfo{author}{Lyman, K.},
  \bibinfo{year}{2017}.
\newblock \bibinfo{title}{Learning to diagnose from scratch by exploiting
  dependencies among labels}.
\newblock \bibinfo{journal}{CoRR} \bibinfo{volume}{abs/1710.10501}.
\bibitem[{Yao et~al.(2018)Yao, Prosky, Poblenz, Covington and
  Lyman}]{DBLP:journals/corr/abs-1803-07703}
\bibinfo{author}{Yao, L.}, \bibinfo{author}{Prosky, J.},
  \bibinfo{author}{Poblenz, E.}, \bibinfo{author}{Covington, B.},
  \bibinfo{author}{Lyman, K.}, \bibinfo{year}{2018}.
\newblock \bibinfo{title}{Weakly supervised medical diagnosis and localization
  from multiple resolutions}.
\newblock \bibinfo{journal}{CoRR} \bibinfo{volume}{abs/1803.07703}.
\newblock \href{http://arxiv.org/abs/1803.07703}{\tt arXiv:1803.07703}.
\bibitem[{Zhang and Sabuncu(2018)}]{noise_mae_loss}
\bibinfo{author}{Zhang, Z.}, \bibinfo{author}{Sabuncu, M.},
  \bibinfo{year}{2018}.
\newblock \bibinfo{title}{Generalized cross entropy loss for training deep
  neural networks with noisy labels}.
\bibitem[{Zhu et~al.(2005)Zhu, Ji, Xu and Gong}]{zhu}
\bibinfo{author}{Zhu, S.}, \bibinfo{author}{Ji, X.}, \bibinfo{author}{Xu, W.},
  \bibinfo{author}{Gong, Y.}, \bibinfo{year}{2005}.
\newblock \bibinfo{title}{Multi-labelled classification using maximum entropy
  method}, pp. \bibinfo{pages}{274--281}.
\newblock \DOIprefix\doi{10.1145/1076034.1076082}.
\bibitem[{Zhu et~al.(2018)Zhu, Liu, Fan and Xie}]{Zhu2018DeepLungD3}
\bibinfo{author}{Zhu, W.}, \bibinfo{author}{Liu, C.}, \bibinfo{author}{Fan,
  W.}, \bibinfo{author}{Xie, X.}, \bibinfo{year}{2018}.
\newblock \bibinfo{title}{Deeplung: Deep 3\uppercase{D} dual path nets for
  automated pulmonary nodule detection and classification}.
\newblock \bibinfo{journal}{2018 IEEE Winter Conference on Applications of
  Computer Vision (WACV)} , \bibinfo{pages}{673--681}.
\bibitem[{Zotin et~al.(2019)Zotin, Hamad, Simonov and
  Kurako}]{lung_boundary_det}
\bibinfo{author}{Zotin, A.}, \bibinfo{author}{Hamad, Y.},
  \bibinfo{author}{Simonov, K.}, \bibinfo{author}{Kurako, M.},
  \bibinfo{year}{2019}.
\newblock \bibinfo{title}{Lung boundary detection for chest {X-ray} images
  classification based on {GLCM} and probabilistic neural networks}.
\newblock \bibinfo{journal}{Procedia Computer Science} \bibinfo{volume}{159},
  \bibinfo{pages}{1439--1448}.
\newblock \DOIprefix\doi{10.1016/j.procs.2019.09.314}.

\end{thebibliography}

\end{document}